\pdfoutput=1

\documentclass[11pt]{article}

\usepackage[final]{acl}

\usepackage{times}
\usepackage{setspace} 
\usepackage{latexsym}
\usepackage{tabularray}  
\usepackage{multirow} 
\usepackage{float}
\usepackage{amssymb}
\usepackage{booktabs}
\usepackage{algorithm}
\usepackage{algpseudocode}
\usepackage{amsmath}
\usepackage{listings}
\usepackage{enumitem}
\usepackage{xcolor}
\usepackage{amsthm}
\theoremstyle{definition}
\newtheorem{definition}{Definition}
\usepackage[T1]{fontenc}

\definecolor{veronica-red}{RGB}{196,30,58}
\definecolor{ForestGreen}{RGB}{34,139,34}
\definecolor{BrickRed}{rgb}{.72,0,0}
\definecolor{LakeBlue}{RGB}{0,61,153}

\usepackage[utf8]{inputenc}
\usepackage{microtype}
\usepackage{inconsolata}
\usepackage{subcaption}
\usepackage{graphicx}

\title{Dynamic and Generalizable Process Reward Modeling}

\author{
Zhangyue Yin\textsuperscript{$\diamondsuit$}\quad 
Qiushi Sun\textsuperscript{$\heartsuit$} \quad 
Zhiyuan Zeng\textsuperscript{$\diamondsuit$} \\
\bf{
Qinyuan Cheng\textsuperscript{$\diamondsuit$} \quad
Xipeng Qiu\textsuperscript{$\diamondsuit$}\textsuperscript{\dag} \quad
Xuanjing Huang\textsuperscript{$\diamondsuit$}\textsuperscript{\dag}
}\\
\textsuperscript{$\diamondsuit$}College of Computer Science and Artificial Intelligence, Fudan University \\
\textsuperscript{$\heartsuit$}The University of Hong Kong \\
\texttt{\{yinzy21,cengzy23,chengqy21\}@m.fudan.edu.cn}\\
\texttt{qiushisun@connect.hku.hk} \quad
\texttt{\{xpqiu,xjhuang\}@fudan.edu.cn}
}
\begin{document}
\maketitle

\begin{abstract}
Process Reward Models (PRMs) are crucial for guiding Large Language Models (LLMs) in complex scenarios by providing dense reward signals. 
However, existing PRMs primarily rely on heuristic approaches,
which struggle with cross-domain generalization.
While LLM-as-judge has been proposed to provide generalized rewards, current research has focused mainly on feedback results, overlooking the meaningful guidance embedded within the text.
Additionally, static and coarse-grained evaluation criteria struggle to adapt to complex process supervision. 
To tackle these challenges, we propose Dynamic and Generalizable Process Reward Modeling (DG-PRM), 
which features a reward tree to capture and store fine-grained, multi-dimensional reward criteria.
DG-PRM dynamically selects reward signals for step-wise reward scoring.
To handle multifaceted reward signals, we pioneeringly adopt Pareto dominance estimation to identify discriminative positive and negative pairs.
Experimental results show that DG-PRM achieves stunning performance on prevailing benchmarks, 
significantly boosting model performance across tasks with dense rewards.
Further analysis reveals that DG-PRM adapts well to out-of-distribution scenarios, demonstrating exceptional generalizability.
\end{abstract}

\begin{flushleft}
\noindent ``Judgements prevent us from seeing the good that lies beyond appearances.'' 
\end{flushleft}
\rightline{--\textit{Wayne Dyer}}

\section{Introduction}
\label{sec:introduction}

\begin{figure}[t]
  \centering
  \includegraphics[width=0.45\textwidth]{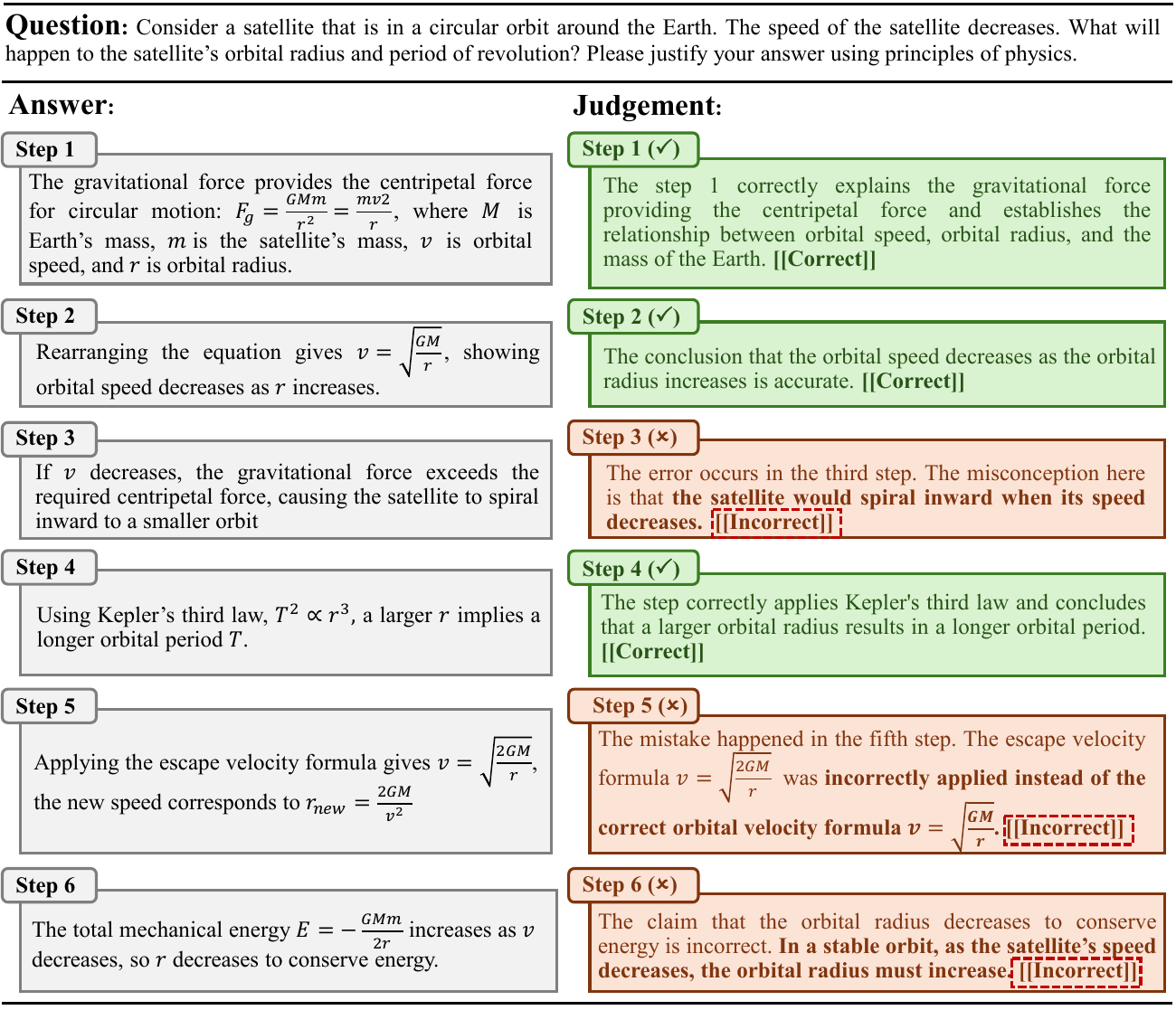}
    \caption{An example from MT-Bench, where LLM judgments contain rich details, as highlighted in bold, but prevailing PRM methods fail to fully exploit.}
  \label{fig:intro}
  \vspace{-1.5em}
\end{figure}

The Process Reward Model (PRM) plays a crucial role in enhancing the performance of Large Language Models (LLMs)~\citep{uesato2023solving,yuan2024freeprocessrewardsprocess,zhang2025lessons}. Unlike outcome reward models (ORM), which provide feedback solely on the final objective, PRM focuses on rewarding the intermediate steps or processes involved in task execution~\citep{setlur2025rewarding}. This approach provides more dense reward signals~\citep{li2024processrewardmodelqvalue}, especially in complex problems where the model needs to reason, analyze, and explore different solution strategies~\citep{wei2022chain}. By evaluating intermediate steps, PRM is essential for improving the model's ability to tackle intricate tasks~\citep{lightman2024lets}.

PRM can be broadly classified into heuristic and generative categories. Heuristic process rewards~\citep{wang2024shepherd,wang2024misp} rely on manually crafted criteria to assess the relevance of intermediate steps to the final answer. While heuristic rewards have significantly advanced the model's reasoning capabilities, they suffer from several limitations~\citep{zhang2025lessons}. 
Specifically, they often require objective, reference-based answers~\citep{luo2024improvemathematicalreasoninglanguage}, which are difficult to obtain in complex scenarios where evaluation criteria fluctuate. 
Furthermore, PRMs trained using heuristic rewards often exhibit poor generalization and even be susceptible to reward hacking~\citep{weng2024rewardhack,wen2025language}. On the other hand, 
generative process rewards utilize LLMs to replace human annotation by labeling each intermediate step as correct or incorrect, offering positive or negative feedback accordingly~\citep{mahan2024generativerewardmodels,cao2024enhancing}. 
While generative rewards capitalize on the LLM's ability to generate responses, the evaluation still relies on a fixed set of standards~\citep{ling2023deductive}, limiting its adaptability across diverse domains.

Furthermore, prevailing LLM-as-Judge methods only utilize final feedbacks (\textit{e.g.}, correct/incorrect) and overlook valuable information encapsulated in the process~\citep{kwon2023reward,gao2023humanlikesummarizationevaluationchatgpt}, such as error severity and the type of mistakes. As illustrated in Figure~\ref{fig:intro}, we observe that LLM feedback contains rich details and guidance information. However, the prevalent approach assigns a uniform negative reward for incorrect labels, 
neglecting the diversity and severity of errors.

In this paper, we identify two key limitations in current process of reward construction: 
(1) the use of fixed evaluation criteria, and 
(2) the reliance on uniform negative rewards, 
which fail to capture the diversity and severity of errors, limiting adaptability and generalizability in process reward.
To address these challenges, we introduce Dynamic and Generalizable Process Reward Modeling (DG-PRM), a novel framework designed to automatically construct and precisely allocate process rewards. 
We propose the use of a reward tree to store multifaceted evaluation criteria extracted from LLM judgments.
It selects the most step-wise relevant criteria during evaluation, thus making DG-PRM excel in cross-domain generalization.
We also introduce Pareto dominance estimation to select positive and negative pairs from a diverse set of reward signals, providing clear optimization objectives. 
Experimental results demonstrate that DG-PRM achieves state-of-the-art performance on \textsc{PRMBench}, showcasing superior PRM capabilities.
By offering contextually appropriate reward signals, 
DG-PRM significantly improves LLM performance across a wide range of tasks. 
Furthermore, compared to the LLM-as-Judge approach, DG-PRM demonstrates enhanced training efficiency and better generalization to out-of-distribution scenarios.

Our main contributions are listed below:
\begin{itemize}[itemsep=2pt,topsep=3pt,parsep=0pt]
    \item We introduce DG-PRM, an automated framework designed to construct dynamic and generalizable process rewards, 
    optimizing the utility of LLM feedback.
    \item To handle diverse and complex rewards, we introduce a novel reward tree to dynamically capture and leverage appropriate criteria to each evaluation step.
    \item We propose the use of Pareto dominance estimation to identify positive and negative pairs from multifaceted reward signals, thereby providing clearer optimization objectives.
    \item DG-PRM significantly boosts LLM performance across a wide range of tasks by offering precise, fine-grained process rewards, while demonstrating high training efficiency and exceptional generalizability.
\end{itemize}

\section{Related Work}
\label{sec:related_work}

\paragraph{Outcome Reward Model.}
Reward models are designed to capture human preferences and automate the evaluation of model outputs~\citep{ouyang2022training, kaufmann2024surveyreinforcementlearninghuman, sun2024improving}. Outcome Reward Model (ORM) has been applied across a broad range of domains, including safety~\citep{dai2024safe}, mathematical problem-solving~\citep{cobbe2021trainingverifierssolvemath, yang2024qwen25mathtechnicalreportmathematical}, and code generation~\citep{dou2024stepcoder, sun2024largelanguagemodeldrivenreward,sun2024neural}. Recent studies, such as \citet{wang2024helpsteer2}, have further expanded the reward signal to encompass diverse dimensions, including helpfulness, correctness, coherence, complexity, and verbosity.
ORM can be categorized into discriminative models~\citep{stiennon2020summarize, ouyang2022training} and generative models~\citep{mahan2024generativerewardmodels}. Discriminative reward models typically add a classification head to assess the quality of inputs~\citep{gao2023scaling, chen2024discriminative}, whereas generative reward models leverage the language generation capabilities of LLMs to evaluate outputs~\citep{zhu2024judgelm, li2024generative}. \citet{zheng2023judging} demonstrate that LLMs can provide scalable and explainable rewards that exhibit high alignment with human preferences.

\paragraph{Process Reward Model.}
As LLMs are increasingly required to handle complex tasks, they often need more tokens to reason effectively~\citep{wei2022chain, snell2024scalingllmtesttimecompute}. As the length of LLM outputs increases, ORM struggles to fully evaluate the coherence and correctness of the output~\citep{luo2024improvemathematicalreasoninglanguage}. 
Thus, building PRMs with dense reward signals emerges as a solution~\citep{lightman2024lets}.
Current approaches of building PRMs tend to focus on objective domains, 
such as mathematics, which have clear, definitive answers~\citep{guan2025rstarmathsmallllmsmaster}, 
aiming to improve LLM performance in mathematical problem-solving~\citep{uesato2022solvingmathwordproblems, yuan2024freeprocessrewardsprocess, cui2025processreinforcementimplicitrewards}. 
\citet{wang2024shepherd} model the correctness of each step’s output as a process reward, 
while \citet{wang2024misp} adopt a softer approach by incorporating the likelihood of the current step’s output being correct as a process reward. However, these heuristic methods limit the generalizability, 
particularly in cases where there is no clear, unique answer. 
For example, in scientific tasks, diverse and complex reward signals should be considered, 
and various output components may require attention to distinct reward signals~\citep{wu2024workflow}. Therefore, constructing dynamic and diverse process rewards is a desideratum.

\paragraph{Reward Signal.}
The design of reward signals is a crucial component in RL~\citep{sutton2018reinforcement}. 
Reward signals can generally be classified into three types: human-annotated~\citep{bai2022training}, rule-based~\citep{glaese2022improvingalignmentdialogueagents}, and AI-feedback~\citep{lee2024rlaifvsrlhfscaling}. 
Human-annotated reward signals require expert labeling and verification, which can be costly and time-consuming~\citep{lightman2024lets}. 
\citet{mu2024rule} argue that human annotations often fail to accurately convey the intended behaviors to annotators, which complicates the conversion of desired outcomes into specific rules. 
Rule-based systems, such as parsing, use compilers to perform syntactic analysis for translating source code into executable binary code.
However, such approaches are domain-specific and cannot easily generalize to other areas. With the increasing capabilities of LLMs, research has shifted toward utilizing AI feedback for reward generation~\citep{bai2022constitutional, bai2023benchmark, li2024llmasajudgesurvey}. \citet{kwon2023reward} observe that LLMs, when used as a proxy reward function, significantly improve the alignment of rewards with user objectives. In the context of code, \citet{mcaleese2024llmcriticshelpcatch} show that AI models help identify more bugs than human contractors. \citet{cao2024enhancing} demonstrate that dense reward signals provided by LLMs effectively improve the performance of policy models. However, current research~\citep{gao2024llmcriticshelpcatch,chen2024improving,ling2023deductive} typically focuses on scoring or ranking outputs, which neglects the rich guiding information present in LLM-generated texts.

\section{Preliminaries}
\label{sec:preliminary}

\subsection{Process Reward Modeling}
Process Reward Models (PRMs) evaluate intermediate steps in model outputs rather than just final outcomes~\citep{lightman2024lets}. Given an input $x$ and model output $\hat{y}$ decomposed into $n$ steps:
\begin{equation}
\hat{y} = \{\hat{y}^{(1)}, \hat{y}^{(2)}, \ldots, \hat{y}^{(n)} | x\}
\end{equation}
PRM assigns a reward signal $r^{(t)}$ to each step $\hat{y}^{(t)}$, providing dense supervision throughout the generation process. 

\subsection{Hierarchical Clustering}
Hierarchical clustering~\citep{johnson1967hierarchical} constructs a tree-like structure of clusters by iteratively merging or splitting based on similarity. Given a set of data points $\{v_1, v_2, \ldots, v_m\}$ in a $d$-dimensional space and a distance function $\mathcal{D}$, hierarchical clustering produces a dendrogram $\mathcal{H}$:
\begin{equation}
\mathcal{H} = \text{HierarchicalCluster}(\{v_1, v_2, \ldots, v_m\}, \mathcal{D})
\end{equation}
Common distance metrics include cosine distance for text embeddings:
\begin{equation}
\mathcal{D}_{\text{cosine}}(v_i, v_j) = 1 - \frac{v_i \cdot v_j}{\|v_i\| \|v_j\|}
\end{equation}

\subsection{Multi-Objective Optimization and Pareto Dominance}
In multi-objective optimization~\citep{miettinen1999nonlinear}, we often face trade-offs between competing objectives. The concept of Pareto dominance provides a framework for comparing solutions:

\begin{definition}[Pareto Dominance]
Given two solutions $y_i$ and $y_j$ evaluated on $k$ objectives with scores $\{s^{(i)}_1, \ldots, s^{(i)}_k\}$ and $\{s^{(j)}_1, \ldots, s^{(j)}_k\}$ respectively, solution $y_i$ \textbf{Pareto-dominates} $y_j$ (denoted $y_i \succ y_j$) if:
\begin{equation}
\forall l \in \{1, \ldots, k\}: s^{(i)}_l \geq s^{(j)}_l \ \text{and} \ \exists l: s^{(i)}_l > s^{(j)}_l
\end{equation}
A solution is \textbf{Pareto-optimal} if no other solution dominates it. The set of all Pareto-optimal solutions forms the \textbf{Pareto front}.
\end{definition}


\section{DG-PRM}
\label{sec:methodology}
\begin{figure*}[t]
  \centering
  \includegraphics[width=0.95\textwidth]{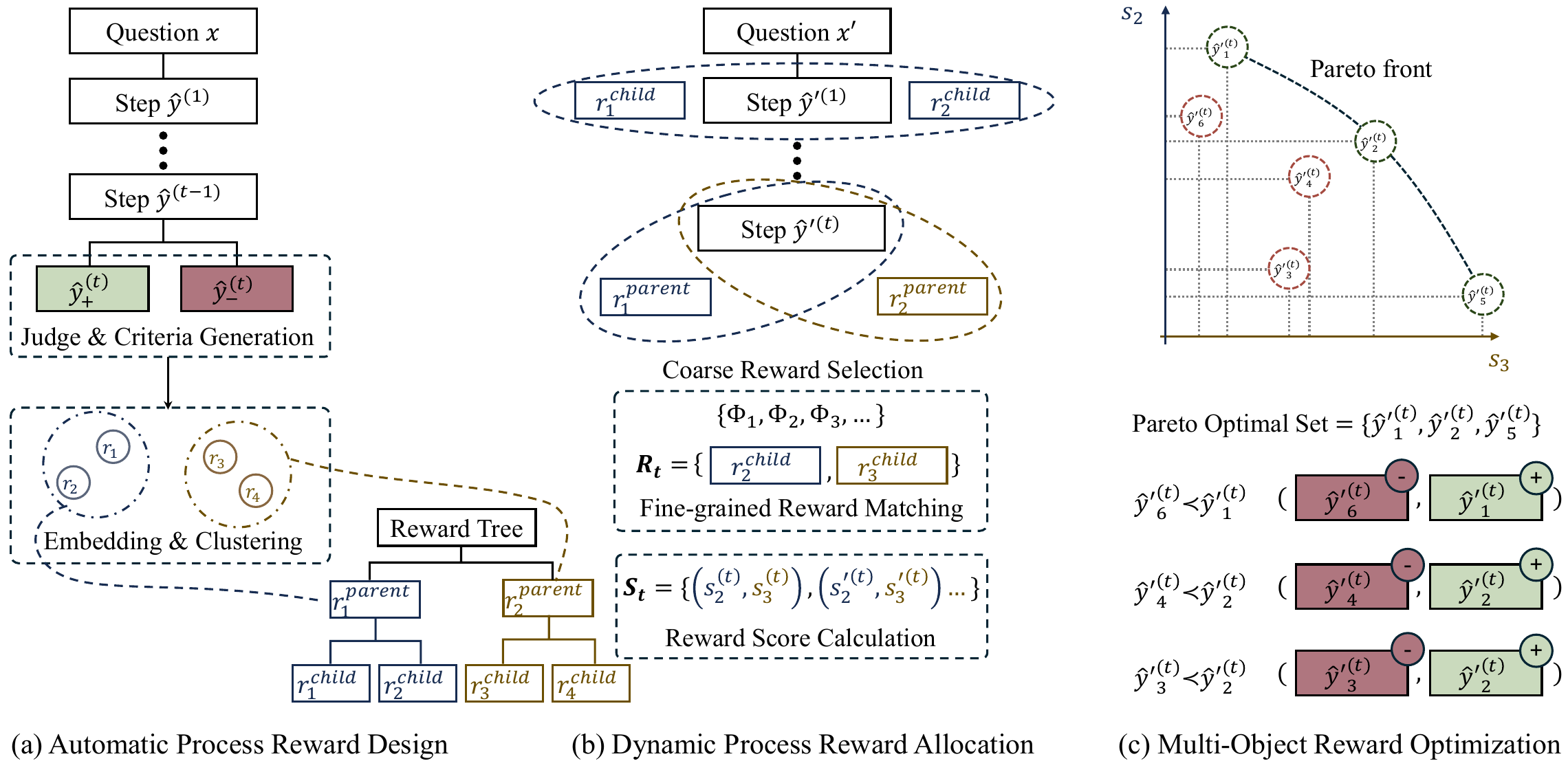}
    \caption{Overview of DG-PRM. DG-PRM consists of three main steps: (a) Automatic Process Reward Design, which constructs reward criteria using positive and negative sample pairs \(\hat{y}_{+}\) and \(\hat{y}_{-}\), maps these criteria into a feature space, and builds a reward tree via hierarchical clustering; 
    (b) Dynamic Process Reward Allocation, which dynamically selects both coarse-grained rewards \(r^{\text{parent}}\) and fine-grained rewards \(r^{\text{child}}\) from the reward tree in each step, computing the reward score based on the corresponding criteria; 
    and (c) Multi-Objective Reward Optimization, which selects the Pareto-optimal \(\hat{y}^{(t)}\) as the optimization target based on the computed reward scores.}  
  \label{fig:overview}
  \vspace{-1em}
\end{figure*}

\begin{algorithm}[t]
\caption{Dynamic Process Reward Allocation}
\begin{algorithmic}[1]
\Require 
     Step-\(t\) \( \hat{y}^{(t)} \), Reward tree \( \mathcal{T} \), Distance threshold \( \zeta \), Window size \( \mu \), Reward function \( \mathcal{R} \), Analysis function \( \Phi \), Embedding function \( \mathcal{V} \), Distance function \( \mathcal{D} \), Score function \( \mathcal{S} \)

\Ensure 
    Reward set \( \mathbf{R} \), Score set \( \mathbf{S} \)

\State \textbf{Initialize} \( \mathbf{R} = \emptyset \), \( \mathbf{S} = \emptyset \)

\For{each timestep \( t = 1 \) to \( n \)}
    \If{\( t - \mu \leq 0 \)}
        \State \textbf{Select} all available previous steps up to \( t-1 \), i.e., \( \{ \hat{y}^{(1)}, \hat{y}^{(2)}, \dots, \hat{y}^{(t-1)} \} \)
    \Else
        \State \textbf{Select} the previous \( \mu \) steps, i.e., \( \{ \hat{y}^{(t-\mu)}, \hat{y}^{(t-\mu+1)}, \dots, \hat{y}^{(t-1)} \} \)
    \EndIf
    
    \State \textbf{Retrieve} corresponding rewards and scores: \( \mathcal{I}_t = \{ (\hat{y}^{(t-\mu)}, r^{(t-\mu)}_k, s_k^{(t-\mu)}), \dots \} \)
    \State \textbf{Add} \( \mathcal{I}_t \) as supplementary information to reward function \( \mathcal{R} \)
    \State \textbf{Cal} \( \phi_i = \Phi(\hat{y}^{(t)}, r^{\text{parent}}_i) \)
    \For{each \( \phi_i \in \mathcal{R}(\hat{y}^{(t)}, \mathcal{T}, \mathcal{I}_t) \)}
        \For{each child \( r^{(t)}_k \) of \( r^{\text{parent}}_i \)}
            \State \( \delta^{(t)}_k = \mathcal{D}(\mathcal{V}(\phi_i), \mathcal{V}(r^{(t)}_k)) \)
            \If{\( \delta^{(t)}_k \le \zeta \)}
                \State \textbf{Add} \( r^{(t)}_k \) to \( \mathbf{R} \)
                \State \textbf{Cal} \( s_k^{(t)} = \mathcal{S}(r^{(t)}_k, \hat{y}^{(t)}, \mathcal{I}_t)\)
                \State \textbf{Add} \( s_k^{(t)} \) to \( \mathbf{S} \)
            \EndIf
        \EndFor
    \EndFor
\EndFor
\end{algorithmic}
\label{alg:dynamic_reward_allocation}
\end{algorithm}


Building on the foundations above, we now present DG-PRM, our novel framework for dynamic and generalizable process reward modeling. As illustrated in Figure~\ref{fig:overview}, the key innovations are: (1) a reward tree structure that captures multi-granular evaluation criteria, (2) a dynamic allocation mechanism that selects contextually appropriate rewards, and (3) Pareto-based optimization for handling diverse reward signals.

\subsection{Automatic Process Reward Design}
\label{subsec:reward_tree}
Unlike existing PRMs that rely on fixed evaluation criteria, DG-PRM automatically extracts diverse reward criteria from comparative analysis. Given positive and negative output pairs $(\hat{y}_+, \hat{y}_-)$, we employ a judge function $\mathcal{J}$ to analyze their differences:
\begin{equation}
R_{\text{raw}} = \bigcup_{(x, \hat{y}_+, \hat{y}_-) \in \mathcal{D}} \mathcal{J}(x, \hat{y}_+, \hat{y}_-),
\end{equation}
where $\mathcal{J}$ outputs a set of reward criteria explaining why step $\hat{y}_+$ is superior to $\hat{y}_-$.

After filtering low-quality criteria through an automated validator (detailed in Appendix~\ref{app:automated_validator}), we obtain a refined set $R = \{r_1, r_2, \ldots, r_m\}$.

To enable efficient retrieval and avoid redundancy, we organize the reward criteria into a hierarchical tree structure. Each criterion $r_i$ is embedded into a $d$-dimensional vector space:
\begin{equation}
v_i = \mathcal{V}(r_i) \in \mathbb{R}^d
\end{equation}

We then apply incremental hierarchical clustering~\citep{zhang1997birch} to construct the reward tree $\mathcal{T}$. To reduce redundancy, criteria with cosine distance below threshold $\xi$ are merged:
\begin{equation}
\text{merge}(r_i, r_j) \quad \text{if} \quad \mathcal{D}_{\text{cosine}}(v_i, v_j) \leq \xi
\end{equation}
The resulting tree structure organizes criteria into coarse-grained parent nodes and fine-grained child nodes:
\begin{equation}
\mathcal{T} = (\{r^{\text{parent}}_1, r^{\text{parent}}_2, \ldots\}, \{r^{\text{child}}_1, \ldots, r^{\text{child}}_m\})
\end{equation}
This hierarchical organization enables efficient navigation from general evaluation aspects to specific criteria.

\subsection{Dynamic Process Reward Allocation}
\label{subsec:dynamic_allocation}
For each step $\hat{y}^{(t)}$ in the model output, DG-PRM dynamically selects relevant rewards from the tree. We first construct the context information from previous steps:
\begin{equation}
\mathcal{I}_t = \{(\hat{y}^{(i)}, r^{(i)}, s^{(i)}) \mid i \in [t-\mu, t-1]\}
\end{equation}
where $\mu$ is the window size for maintaining computational efficiency.

The reward function $\mathcal{R}$ identifies appropriate parent criteria from the reward tree:
\begin{equation}
\{r^{\text{parent}}_1, r^{\text{parent}}_2, \ldots\} = \mathcal{R}(\hat{y}^{(t)}, \mathcal{T}, \mathcal{I}_t)
\end{equation}

For each parent criterion $r^{\text{parent}}_i$, an analysis function $\Phi$ determines whether fine-grained evaluation is needed and generates corresponding evaluation aspects:
\begin{equation}
\{\phi_{i,1}, \phi_{i,2}, \ldots\} = \Phi(\hat{y}^{(t)}, r^{\text{parent}}_i)
\end{equation}
where an empty set indicates that only coarse-grained evaluation is sufficient.

When fine-grained evaluation aspects are generated, we match them to child nodes. For each evaluation aspect $\phi_{i,j}$ and each child node $r_k \in \text{children}(r^{\text{parent}}_i)$, we compute:
\begin{equation}
\delta_k = \mathcal{D}_{\text{cosine}}(\mathcal{V}(\phi_{i,j}), \mathcal{V}(r_k))
\end{equation}

A child criterion $r_k$ is selected if its distance to any evaluation aspect is below the threshold $\zeta$:
\begin{equation}
r_k \in \mathbf{R}_t \iff \exists \phi_{i,j} : \delta_k \leq \zeta
\end{equation}


The final reward set for step $t$ combines all selected criteria:
\begin{equation}
\mathbf{R}_t = \bigcup_{i \in \mathcal{I}} \{r_k \mid r_k \text{ selected from } r^{\text{parent}}_i\}
\end{equation}
where $\mathcal{I} = \{i \mid r^{\text{parent}}_i \text{ was selected by } \mathcal{R}\}$.

Each selected reward $r_k \in \mathbf{R}_t$ is then scored considering the current step and context:
\begin{equation}
s_k^{(t)} = \mathcal{S}(r_k, \hat{y}^{(t)}, \mathcal{I}_t)
\end{equation}

Algorithm~\ref{alg:dynamic_reward_allocation} summarizes the dynamic reward allocation process, which forms the core of DG-PRM's ability to provide contextually appropriate process rewards.

\subsection{Multi-Objective Reward Optimization}
\label{subsec:pareto_optimization}

Given the diverse reward signals in $\mathbf{R}_t$, we employ Pareto dominance to identify clear optimization targets. For multiple candidate outputs $\{\hat{y}^{(t)}_1, \hat{y}^{(t)}_2, \ldots\}$ at step $t$, we compute their Pareto-optimal set:
\begin{equation}
\mathbf{U} = \{\hat{y}^{(t)}_i \mid \nexists \hat{y}^{(t)}_j : \hat{y}^{(t)}_j \succ \hat{y}^{(t)}_i\}
\end{equation}
From this, we construct preference pairs where Pareto-optimal solutions are preferred over dominated ones:
\begin{equation}
\mathbf{V} = \{(\hat{y}^{(t)}_+, \hat{y}^{(t)}_-) \mid \hat{y}^{(t)}_+ \in \mathbf{U}, \exists \hat{y}^{(t)}_- : \hat{y}^{(t)}_+ \succ \hat{y}^{(t)}_-\}
\end{equation}
We adapt DPO~\citep{rafailov2023direct} for step-wise optimization with context dependency:
\begin{equation}
\mathcal{L}_{\text{DG-PRM}}(\theta) = -\mathbb{E}_{(\hat{y}_+^{(t)}, \hat{y}_-^{(t)}) \in \mathbf{V}} \left[ \log \sigma \left( \beta \Delta^{(t)} \right) \right]
\end{equation}
where $\Delta^{(t)}$ measures the log-ratio difference between preferred and dispreferred outputs:
\begin{align}
\Delta^{(t)} &= r_\theta^{(t)}(\hat{y}_+^{(t)}) - r_\theta^{(t)}(\hat{y}_-^{(t)}) \\
r_\theta^{(t)}(\hat{y}) &= \log \frac{\pi_\theta(\hat{y}|x, \hat{y}^{(<t)})}{\pi_{\text{ref}}(\hat{y}|x, \hat{y}^{(<t)})}
\end{align}
Here, $r_\theta^{(t)}(\hat{y})$ represents the log-ratio of the policy $\pi_\theta$ relative to the reference policy $\pi_{\text{ref}}$ for generating step $\hat{y}$ given the input $x$ and previous steps $\hat{y}^{(<t)}$. 
This formulation ensures that the model learns to generate steps that are Pareto-optimal with respect to the dynamically selected reward criteria.

\section{Experiments}
\label{sec:experiment}
\begin{table*}[t]
\belowrulesep=0pt
\aboverulesep=0pt
\fontsize{14}{21}\selectfont
\centering

\resizebox{\textwidth}{!}{
\begin{tabular}{l c|ccc| ccccc| cc cc}
\toprule[1.5pt]
\multirow{2}{*}{\textbf{Model Name}}& \multirow{2}{*}{\textbf{Overall}}  & \multicolumn{3}{c|}{\textbf{Simplicity}}  & \multicolumn{5}{c|}{\textbf{Soundness}}& \multicolumn{4}{c}{\textbf{Sensitivity}}\\
\cmidrule(lr){3-5} \cmidrule(lr){6-10} \cmidrule(lr){11-14} 
&& \textbf{NR.} & \textbf{NCL.} & \textbf{Avg.} &\textbf{ES} &\textbf{SC.}&\textbf{DC.} &\textbf{CI} & \textbf{Avg.} &\textbf{PS} & \textbf{DR.} & \textbf{MS.} & \textbf{Avg.}   \\
\midrule
\hline \multicolumn{14}{c}{\textit{\textbf{Open-source Discriminative Process Reward Model}}} \\   \hline 
\href{https://huggingface.co/ScalableMath/llemma-7b-prm-prm800k-level-1to3-hf}{Llemma-PRM800k-7B$^\dagger$} & 52.0 & 49.3 & 53.4 & 51.4 & 56.4 & 47.1 & 46.7 & 53.3 & 50.9 & 51.0 & 53.5 & 93.6 & 66.0\\
\href{https://github.com/KbsdJames/MATH-Minos}{MATHMinos-Mistral-7B$^\dagger$} & 54.2 & 48.8 & 54.0 & 51.4 & 57.0 & 52.1 & 50.7 & 57.8 & 54.4 & 52.8 & 55.8 & 91.1 & 66.5\\
\href{https://huggingface.co/peiyi9979/math-shepherd-mistral-7b-prm}{MathShepherd-Mistral-7B$^\dagger$} & 47.0 & 44.0 & 50.3 & 47.1 & 49.4 & 44.5 & 41.3 & 47.7 & 45.7 & 47.2 & 48.6 & 86.1 & 60.7\\
\href{https://huggingface.co/RLHFlow/Llama3.1-8B-PRM-Mistral-Data}{RLHFlow-PRM-Mistral-8B$^\dagger$} & 54.4 & 46.1 & 47.3 & 46.7 & 56.6 & 55.1 & 54.4 & 63.8 & 57.5 & 51.5 & 56.2 & 97.9 & 68.5\\
\href{https://huggingface.co/RLHFlow/Llama3.1-8B-PRM-Deepseek-Data}{RLHFlow-PRM-Deepseek-8B$^\dagger$} & 54.2 & 46.4 & 48.9 & 47.6 & 55.7 & 55.0 & 53.2 & 66.2 & 57.5 & 49.0 & 55.4 & 99.8 & 68.1\\

\hline \multicolumn{14}{c}{\textit{\textbf{Prompted as Critic Models}}} \\   \hline 
\href{https://openai.com/index/openai-o1-mini-advancing-cost-efficient-reasoning/}{o1-mini$^{\dagger*}$} & 68.8 & 65.6 & 63.7 & 64.6 & 74.5 & 67.7 & 73.8 & 72.3 & 72.1 & 61.8 & 64.8 & \textbf{100.0} & 75.5\\
\href{https://openai.com/index/hello-gpt-4o/}{GPT-4o$^\dagger$} & 66.8 & 57.0 & 62.4 & 59.7 & 72.0 & 69.7 & 70.7 & 71.1 & 70.9 & 62.5 & 65.7 & 99.2 & 75.8\\
\href{https://huggingface.co/Qwen/QwQ-32B-Preview/}{QwQ-Preview-32B$^\dagger$} & 63.6 & 57.2 & 55.6 & 56.4 & 67.4 & 72.3 & 66.2 & 66.9 & 68.2 & 57.8 & 62.7 & \textbf{100.0} & 73.5\\
\href{https://huggingface.co/deepseek-ai/DeepSeek-R1-Distill-Qwen-32B/}{R1-Distill-Qwen-32B$^\dagger$} & 60.2 & 57.2 & 51.9 & 54.5 & 66.1 & 68.4 & 69.3 & 64.8 & 67.2 & 53.3 & 54.6 & 99.9 & 69.3\\
\href{https://huggingface.co/deepseek-ai/DeepSeek-R1-Distill-Qwen-7B/}{R1-Distill-Qwen-7B$^\dagger$} & 52.6 & 32.9 & 37.9 & 35.4 & 47.3 & 54.1 & 48.4 & 48.0 & 49.4 & 45.6 & 46.8 & \textbf{100.0} & 64.1\\
\href{https://huggingface.co/deepseek-ai/DeepSeek-R1/}{DeepSeek-R1} & 69.5 & 66.0 & 65.2 & 65.6 & 74.8 & 70.1 & 72.2 & 72.9 & 72.5 & 63.2 & 66.2 & \textbf{100.0} & 76.5 \\

\hline \multicolumn{14}{c}{\textit{\textbf{Dynamic and Generalizable Process Reward Modeling}}} \\   \hline 
\href{https://openai.com/index/openai-o1-mini-advancing-cost-efficient-reasoning/}{o1-mini} & 73.5 & 71.2 & 69.1 & 70.2 & 77.5 & 74.8 & 76.3 & 75.6 & 76.1 & 67.3 & 70.4 & \textbf{100.0} & 79.2 \\
\href{https://openai.com/index/hello-gpt-4o/}{GPT-4o} & 72.3 & 66.1 & 69.0 & 67.6 & 75.9 & 73.2 & 74.7 & 76.4 & 75.1 & 66.8 & 70.9 & \textbf{100.0} & 79.2 \\
\href{https://huggingface.co/Qwen/QwQ-32B-Preview/}{QwQ-Preview-32B} & 70.0 & 63.2 & 65.4 & 64.3 & 72.4 & 74.3 & 72.9 & 74.5 & 73.5 & 64.5 & 67.9 & \textbf{100.0} & 77.5 \\
\href{https://huggingface.co/deepseek-ai/DeepSeek-R1-Distill-Qwen-32B/}{R1-Distill-Qwen-32B} & 69.0 & 62.0 & 64.7 & 63.4 & 71.1 & 72.6 & 71.3 & 73.8 & 72.2 & 63.6 & 66.8 & \textbf{100.0} & 76.8 \\
\href{https://huggingface.co/deepseek-ai/DeepSeek-R1-Distill-Qwen-7B/}{R1-Distill-Qwen-7B} & 65.2 & 60.4 & 62.1 & 61.3 & 69.8 & 68.1 & 69.5 & 72.1 & 69.9 & 62.4 & 64.2 & \textbf{100.0} & 75.5 \\
\href{https://huggingface.co/deepseek-ai/DeepSeek-R1/}{DeepSeek-R1} & \textbf{76.5} & \textbf{74.1} & \textbf{72.3} & \textbf{73.2} & \textbf{80.1} & \textbf{77.5} & \textbf{78.9} & \textbf{79.4} & \textbf{79.0} & \textbf{71.0} & \textbf{74.3} & \textbf{100.0} & \textbf{81.8} \\
   
\bottomrule[1.5pt]
\end{tabular}}
\caption{
Performance comparison of DG-PRM and other strong baselines on \textsc{PRMBench}~\citep{song2025prmbench} ($PRM$-Score \%). 
The best results are highlighted in \textbf{bold}. $^{\dagger}$ indicates results from the \href{https://prmbench.github.io/}{official leaderboard}, and $^{*}$ denotes evaluation on a subset of 394 samples. The evaluation includes mainstream open-source and closed-source models for a fair comparison. Details of each dataset category and evaluation objectives are provided in Appendix~\ref{app:dataset_details}.
}
\vspace{-1em}
\label{tab:main_results}
\end{table*}

\begin{figure*}[t]
  \centering
  \begin{subfigure}{0.485\linewidth}
    \includegraphics[width=\linewidth]{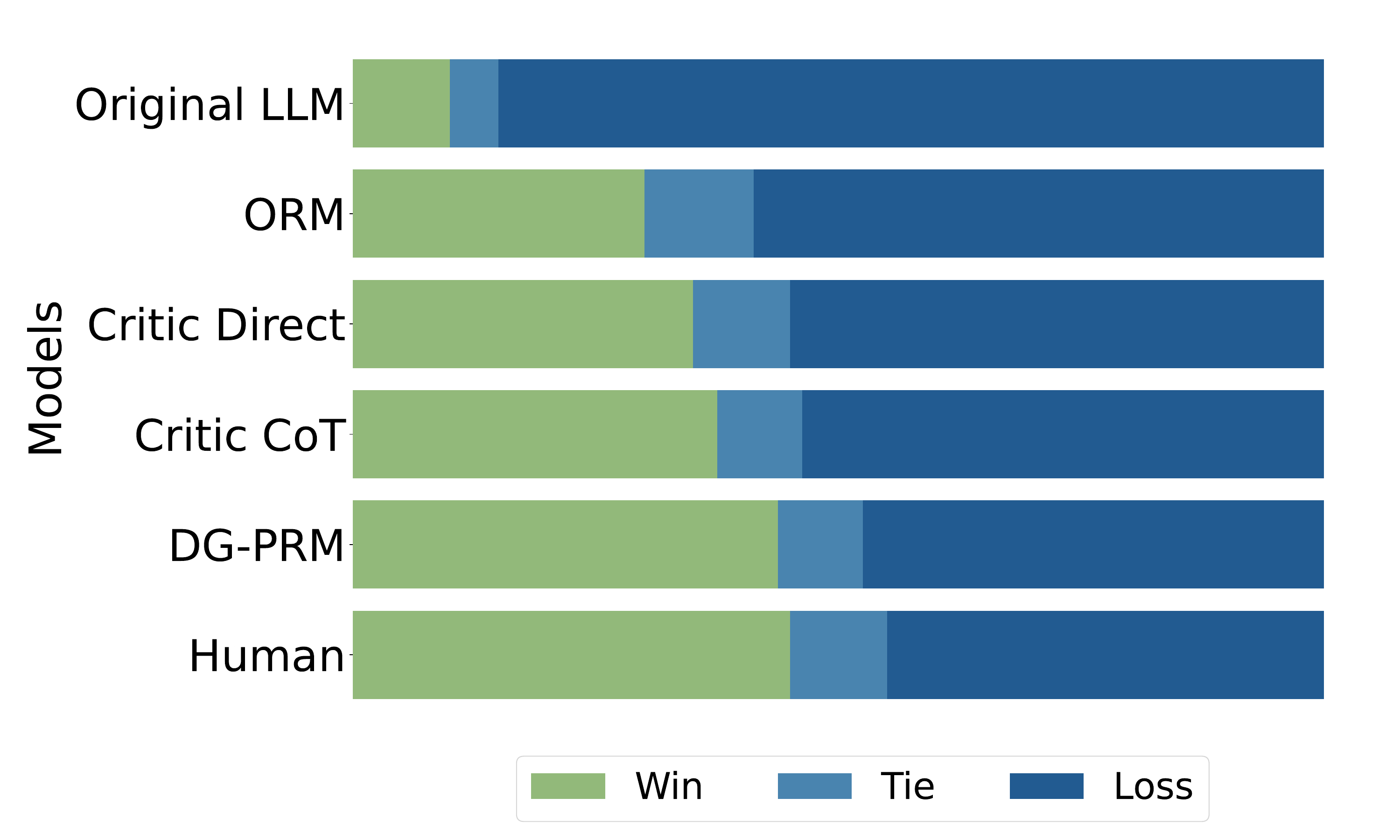}
    \caption{MT-Bench}
    \label{fig:mtbench_win_rate}
  \end{subfigure}
  \hfill 
  \begin{subfigure}{0.49\linewidth}
    \raisebox{0.5ex}{\includegraphics[width=\linewidth]{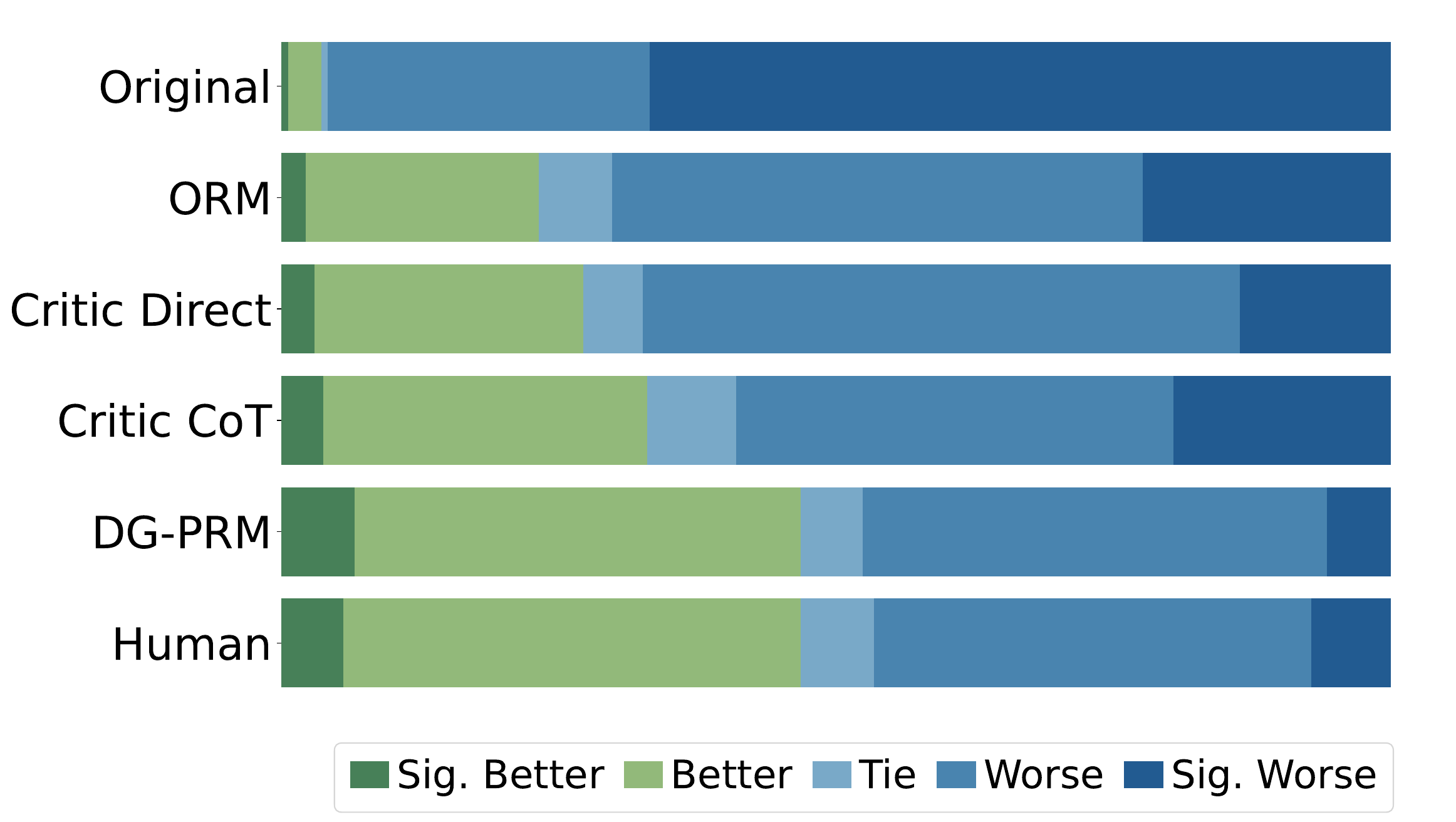}}
    \caption{Arena-Hard}
    \label{fig:arean_hard_win_rate}
  \end{subfigure}
    \caption{Performance comparison on (a) MT-Bench and (b) Arena-Hard. The R1-Distill-Qwen-7B model is used as the backbone, with GPT-4o~\citep{openai2024gpt4o} serving as the judge model.}
\vspace{-1em}
\label{fig:win_rate}
\end{figure*}

\subsection{Evaluation Datasets}
\label{subsec:evaluation_datasets}

To evaluate the efficiency of DG-PRM in process reward modeling, we use the \textsc{PRMBench}~\citep{song2025prmbench} dataset. This benchmark, built on the PRM800K corpus, consists of 6k math problems across 9 distinct error categories, enabling comprehensive evaluation of process-level reward models.

To further assess DG-PRM's effectiveness across a broad range of tasks, we incorporate three additional task sets representing different domains: general tasks, scientific tasks, and commonsense reasoning. These datasets include:

\begin{itemize}[itemsep=2pt,topsep=3pt,parsep=0pt]
    \item \textit{General}: MT-Bench~\citep{zheng2023judging}, Arena-Hard~\citep{arenahard2024}
    \item \textit{Science}: QASC~\citep{khot2020qasc}, ChemistryQA~\citep{wei2021chemistryqa}
    \item \textit{Commonsense}: StrategyQA~\citep{geva2021strategyqa}, ARC-c~\citep{clark2018think}
\end{itemize}

A detailed description of the datasets, including answer types, training and test set distributions, and licensing information, can be found in Appendix~\ref{app:dataset_details}.

\subsection{Experiment Settings}
\label{subsec:experiment_settings}

\paragraph{Implementation Details.}  
For \textsc{PRMBench}, we use the same setup as in the \textit{Prompted as Critic Models} configuration, which includes four open-source models: QwQ-Preview-32B~\citep{qwq-32b-preview}, 
DeepSeek-R1-Distill-Qwen-32B, DeepSeek-R1-Distill-Qwen-7B, and DeepSeek-R1~\citep{guo2025deepseek}, along with two proprietary models: o1-mini~\citep{openai2024o1mini} and GPT-4o~\citep{openai2024gpt4o}. 
In \textit{General}, \textit{Science}, and \textit{Commonsense} scenarios, 
we also include the models DeepSeek-R1-Distill-Qwen-1.5B and DeepSeek-R1-Distill-Qwen-14B. To construct the reward tree, we use the BAAI/bge-en-icl model~\citep{li2024makingtextembeddersfewshot} to build \(\mathcal{V}\), with the dimensionality \(d\) set to 4096. For hierarchical clustering, we apply the BIRCH algorithm~\citep{zhang1997birch} to create \(\mathcal{H}\). Unless otherwise stated, we use DeepSeek-R1-Distill-Qwen-7B as the backbone model and set the hyperparameters as follows: the merge hyperparameter \(\xi = 0.25\), the distance hyperparameter \(\zeta = 0.2\), and step hyperparameter \(\mu = 20\). For more detailed experimental procedures and hyperparameter analysis, refer to Appendix~\ref{app:implement_details} and Appendix~\ref{app:hyperparameter_analysis}.

\paragraph{Baselines.} For \textsc{PRMBench}, we use the official results as baselines.
To further validate the effectiveness of DG-PRM,
we compare it against the following setups in the \textit{General}, \textit{Science}, and \textit{Commonsense} scenarios:
\textbf{(1) Original:} The original model without any optimization.
\textbf{(2) ORM:} We train a reward model to provide positive and negative feedback for complete outputs.
\textbf{(3) Critic Models:} Building on \citet{song2025prmbench}, we guide LLMs with prompts to critique the solution in a step-by-step manner. Following \citet{mahan2024generativerewardmodels}, we distinguish between \textbf{Critic Direct}, where models directly generate results, and \textbf{Critic CoT}, where the model first performs reasoning and analysis before presenting the answer.
\textbf{(4) Human Annotation:} Training on human-annotated process labels as an upper bound.

We also include \textbf{Implicit PRM}~\citep{yuan2024freeprocessrewardsprocess}, a token-level reward baseline in our analysis, as detailed in Appendix~\ref{app:implicit_prm}.
We obverse that a substantial amount of preference data is essential for training in order to effectively model DPO-equivalent rewards~\citep{rafailov2024from}. 

\subsection{Main Results}
\label{subsec:main_results}
\paragraph{\textsc{PRMBench}.}
Table~\ref{tab:main_results} presents the results on \textsc{PRMBench}, where generative models demonstrate a clear advantage over discriminative models, a finding consistent with \citet{zheng2024processbenchidentifyingprocesserrors}. DG-PRM further significantly improves the process reward modeling capabilities of LLMs. In more challenging PS tasks, where PRMs must be capable of identifying missing conditions and prerequisite mistakes, R1-Distill-Qwen-7B shows a remarkable improvement from 45.6\% to 62.4\%. This notable enhancement can be attributed to DG-PRM’s stepwise discrimination and dynamic reward allocation, which effectively increase error identification accuracy. 
Moreover,
R1 demonstrates strong performance, which DG-PRM further enhances, improving overall accuracy from 69.5\% to 76.5\% as a new state-of-the-art.
Notably, 
in sensitivity testing, the average score exceeds 80\%. This can be attributed to R1’s complex reasoning generating higher-quality reward criteria, which DG-PRM effectively leverages to boost performance.
We will further analyze this phenomenon in Appendix~\ref{app:judge_model}.

\vspace{-.5em}
\paragraph{General.} Table~\ref{fig:win_rate} presents the results on MT-Bench and Arean-Hard.
We observe that preference optimization significantly enhances the model's win rate. The Critical Model outperforms ORM through step-wise analysis, achieving better performance. 
In Arean-Hard, using Critical CoT with process analysis demonstrates a clear advantage over Critical Direct, highlighting the importance of step-wise analysis during evaluation. DG-PRM notably improves the model's win rate compared to the baseline, reaching levels close to Human Annotation, and even exceeding Human Annotation in the ``Significantly Better'' category on Arean-Hard. 
We attribute this to DG-PRM's ability to integrate multi-dimensional reward information, resulting in more accurate process rewards.

\begin{figure}
    \centering
    \includegraphics[width=\linewidth]{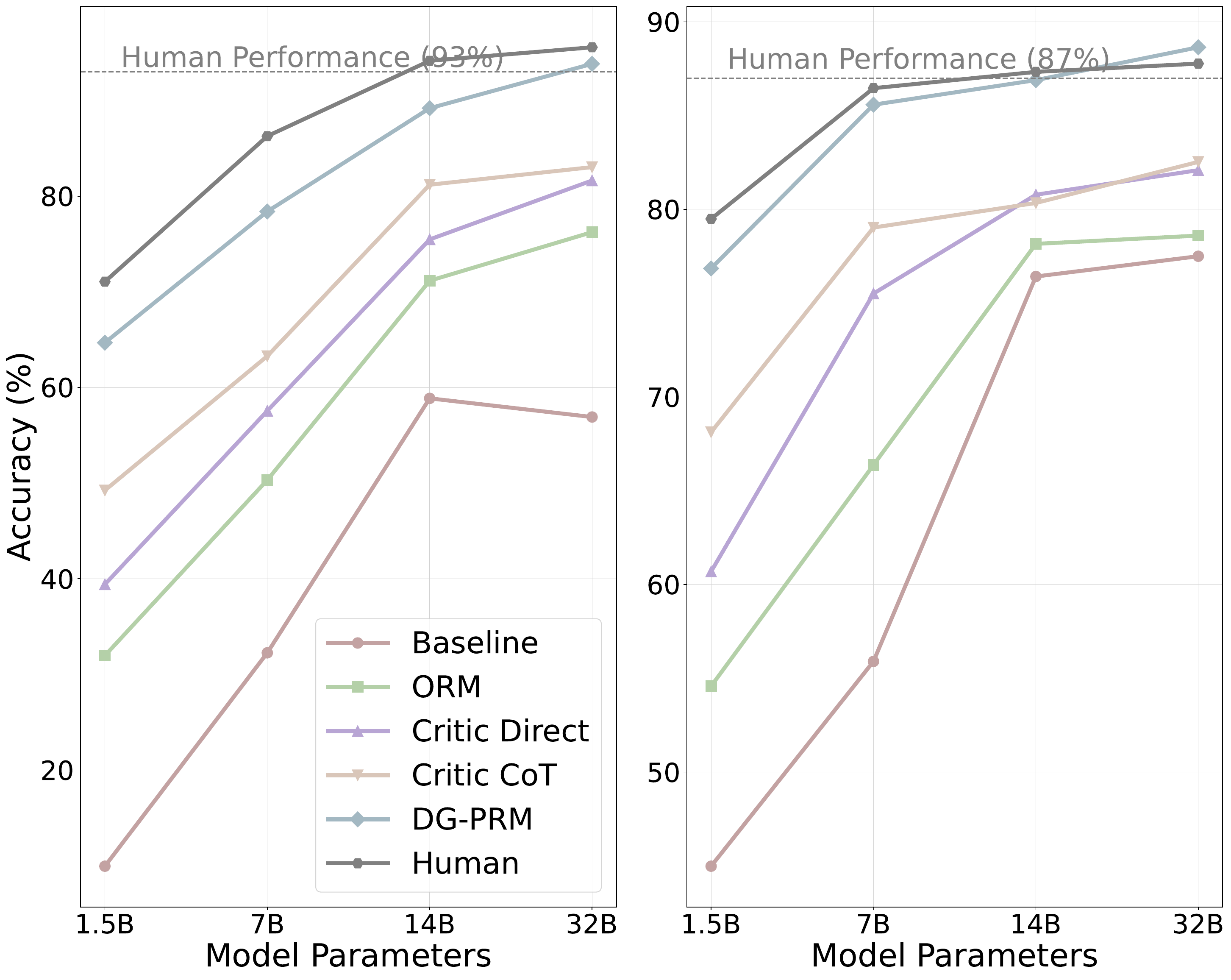}
    \caption{Performance comparison on (a) QASC and (b) StrategyQA. The evaluation includes models of different parameter scales: 1.5B, 7B, 14B, and 32B.}
    \label{fig:parameters}
    \vspace{-1.25em}
\end{figure}

\begin{figure*}[t]
  \centering
  \begin{subfigure}{0.49\linewidth}
    \includegraphics[width=\linewidth]{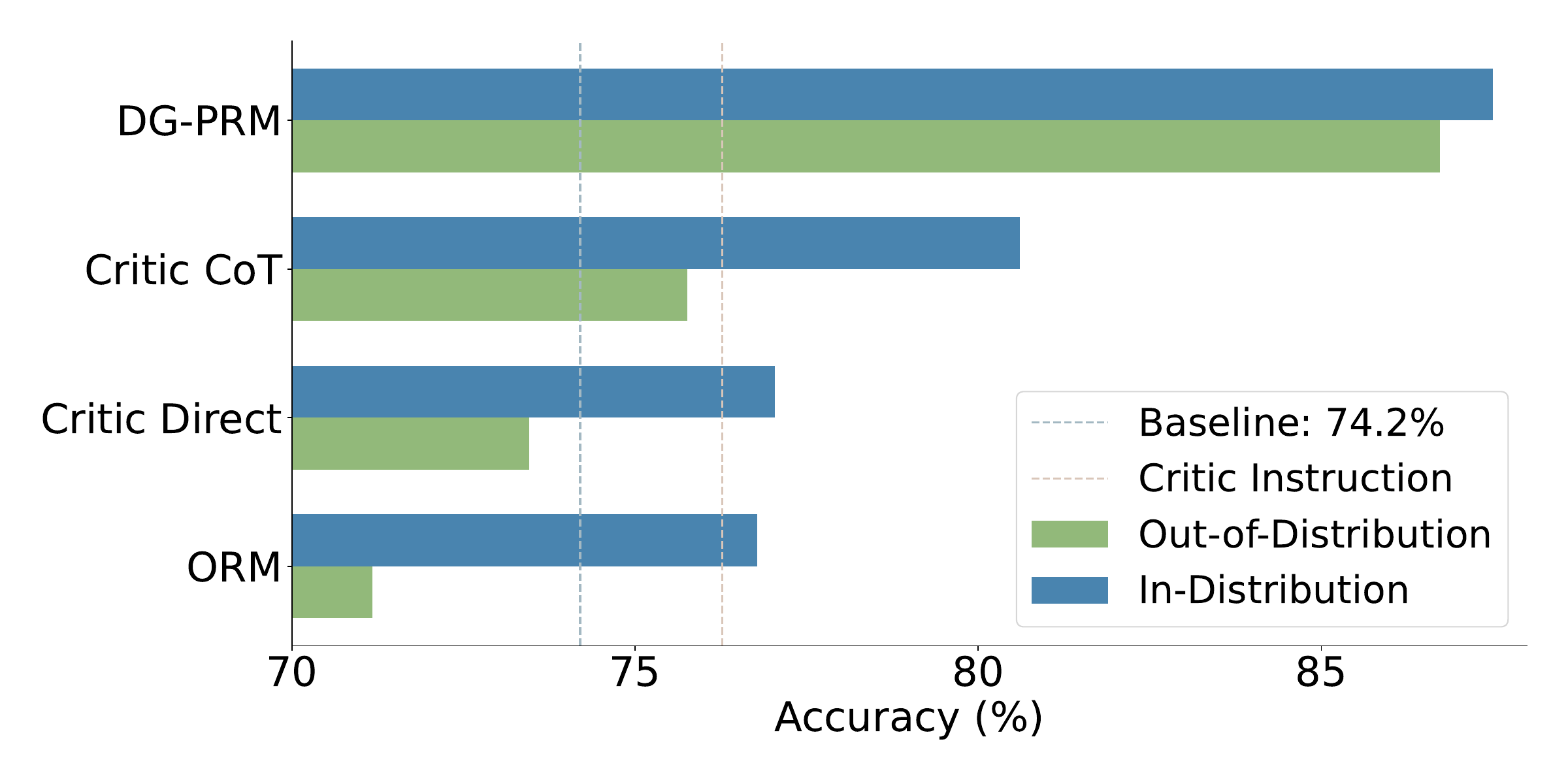}
    \caption{ChemistryQA}
    \label{fig:chemistryqa_generalization}
  \end{subfigure}
  \hfill 
  \begin{subfigure}{0.49\linewidth}
    \includegraphics[width=\linewidth]{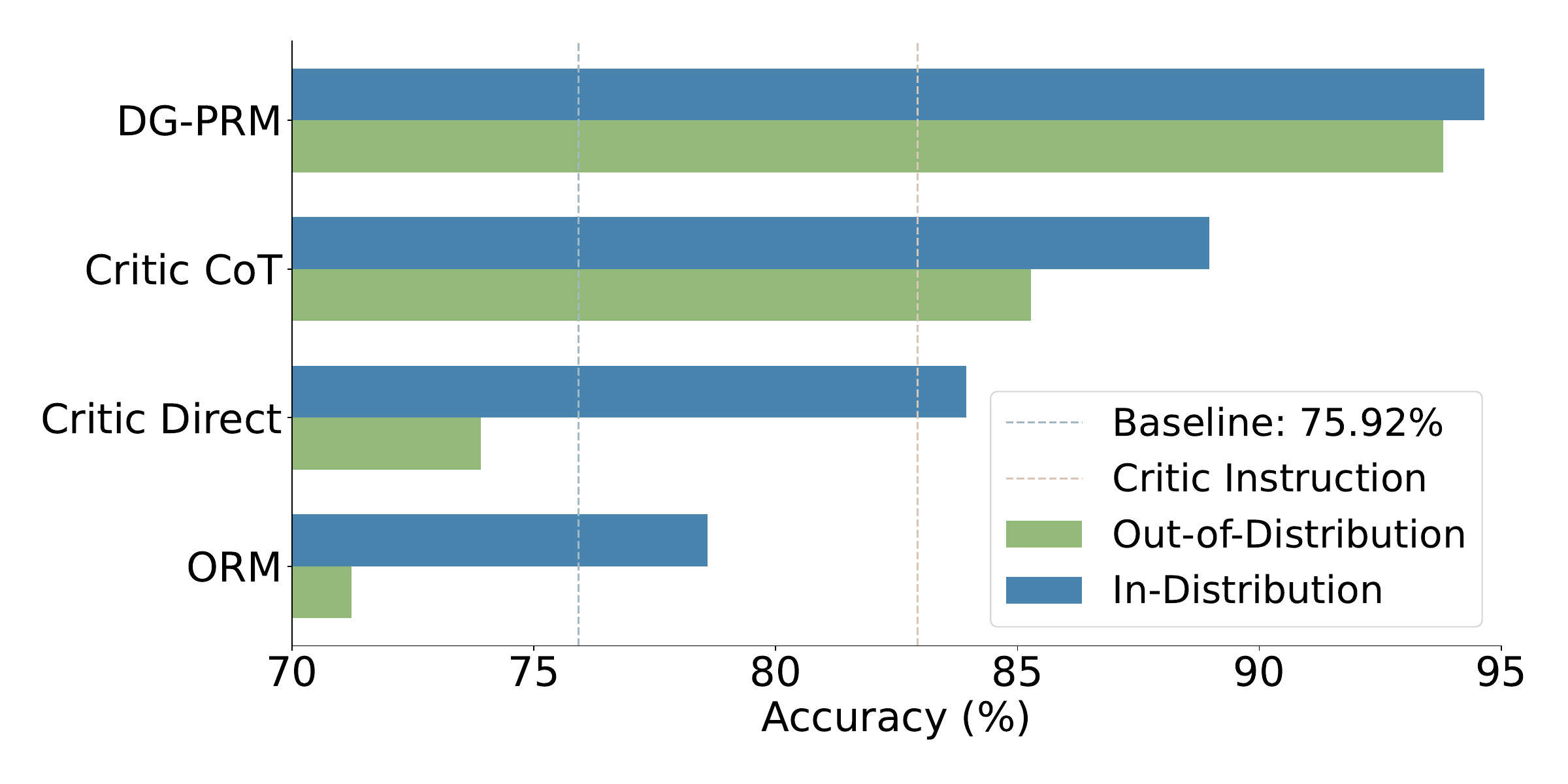}
    \caption{ARC-c}
    \label{fig:arc_challenge_generalization}
  \end{subfigure}
    \caption{Generalization analysis on (a) ChemistryQA and (b) ARC-c. In the Out-of-Distribution setting, we provide process feedback for ChemistryQA samples using the process reward model constructed on QASC, and for ARC-c samples using the process reward model constructed on StrategyQA. \textbf{Critic-Instruction} refers to the approach where only instructions are used without providing any domain-specific exemplars.}
\vspace{-1em}
\label{fig:generalization}
\end{figure*}

\begin{figure*}[t]
  \centering
  \includegraphics[width=\linewidth]{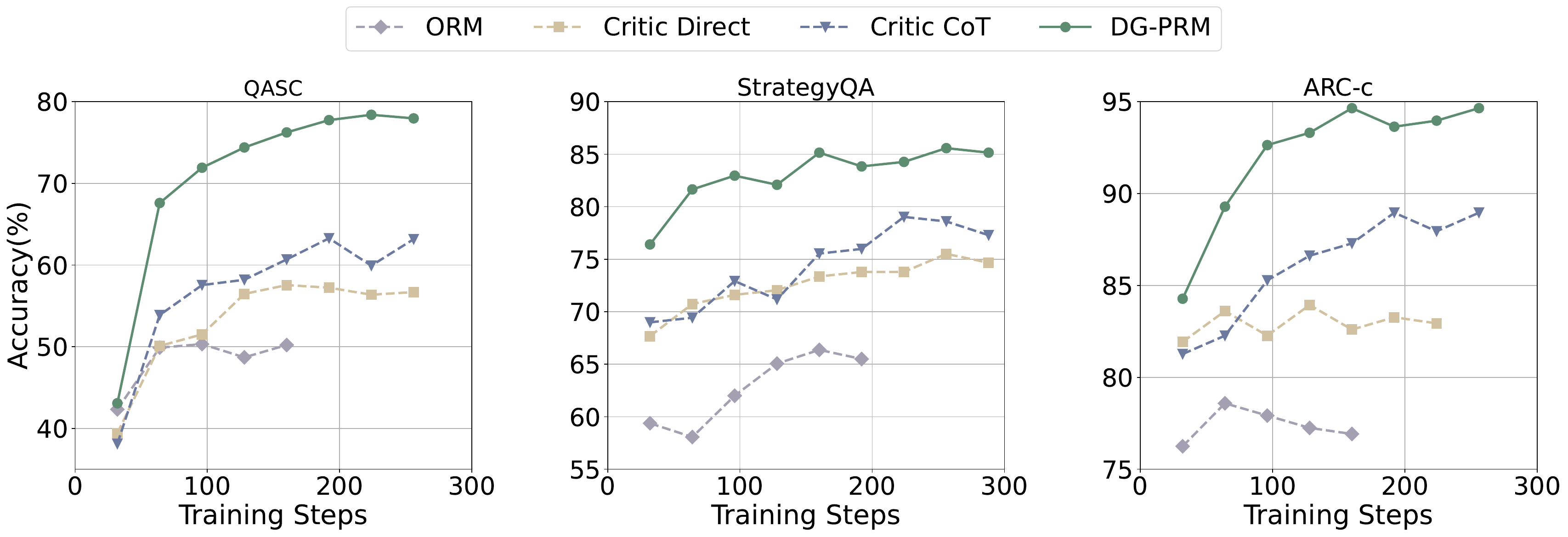}
    \caption{Accuracy($\%$) variation with training steps on the QASC, StrategyQA, and ARC-c datasets.}
  \label{fig:learning_curve}
  \vspace{-1em}
\end{figure*}
\begin{table*}[tbp]
\centering
\begin{tabular}{lcccc}
\toprule
& \textbf{QASC} & \textbf{ChemistryQA} & \textbf{StrategyQA} & \textbf{ARC-c} \\
\midrule
\# Coarse-grained Reward    & 2.3  & 0.9  & 1.5  & 1.2  \\
\# Fine-grained Reward      & 1.3  & 3.7  & 2.1  & 1.9  \\
\midrule
Selection Ratios            & 0.81 & 0.76 & 0.62 & 0.79 \\
w/o Pareto-based filtering  & 75.70\% & 85.71\% & 82.97\% & 90.63\% \\
w. Pareto-based filtering   & 78.40\%{\color{red} ($\uparrow$2.70\%)} & 87.50\%{\color{red} ($\uparrow$1.79\%)} & 85.58\%{\color{red} ($\uparrow$2.61\%)} & 94.65\%{\color{red} ($\uparrow$4.02\%)} \\
\bottomrule
\end{tabular}
\caption{ Statistical analysis and ablation study of DG-PRM across datasets, where \# denotes the number of rewards.}
\label{tab:ablation_comparison}
\vspace{-1em}
\end{table*}

\vspace{-0.5em}
\paragraph{Science and Commonsense.} 
In Figure~\ref{fig:parameters}, we compare the performance of different methods across models with 1.5B, 7B, 14B, and 32B parameters. We observe that DG-PRM significantly enhances the performance of models at all parameter scales, demonstrating a clear advantage over other methods. 
Furthermore, DG-PRM achieves performance close to human annotation, even surpassing human-level performance on the StrategyQA.
Notably, the R1-Distill-Qwen-32B model trained with DG-PRM outperforms human performance on both the QASC and StrategyQA datasets. In StrategyQA, 
we observe a performance drop for the Critical CoT method,
which even falls below the Critical Direct method in the 14B model.
This is due to the implicit reasoning challenges in StrategyQA, where the model often struggles to determine the appropriate direction for analysis, particularly when the reasoning process is highly complex. 
In such cases, unoriented analysis can lead to erroneous judgments. DG-PRM, by providing relevant and explicit reward objectives at each step, effectively improves the judgments.

\vspace{-0.5em}
\paragraph{Generalization.} In Figure~\ref{fig:generalization}, we analyze the generalization capabilities of different methods. We observe a significant performance drop for ORM and Critic Direct in out-of-domain (OOD) scenarios, with some results falling below baseline performance. To further analyze the Critical methods, we introduce Critic Instruction, which provides no examples and only simple instructions to guide the LLM in generating process rewards. In the OOD setting, the Critical CoT method exhibits minimal performance degradation on the ARC-c, but a significant drop on ChemistryQA, even performing worse than Critic Instruction. This is attributed to the diverse problem distribution in QASC, which leads to confusion when transferring to the Chemistry domain. 
In contrast, DG-PRM, by selecting relevant and effective rewards, demonstrates outstanding domain generalization abilities. In Appendix~\ref{app:general_reward_tree}, we leverage the scalability of DG-PRM to construct a general reward tree for further analysis of its cross-domain generalization.

\vspace{-0.5em}
\paragraph{Training Efficiency.} 
In Figure~\ref{fig:learning_curve}, we compare the accuracy of different methods as a function of training steps. We observe that DG-PRM demonstrates exceptional training efficiency across all datasets, achieving performance equivalent to Critic CoT with only 30\% of the training steps. 
This efficiency is attributed to DG-PRM's use of Pareto advantage estimation to select the most representative positive and negative samples, making it easier for the model to capture the differences between samples and optimize the fitting objective. 
As a result, DG-PRM continuously improves model performance, significantly outperforming baseline methods.

\subsection{Analysis}
\label{subsec:analysis}

\begin{figure}
    \centering
    \includegraphics[width=\linewidth]{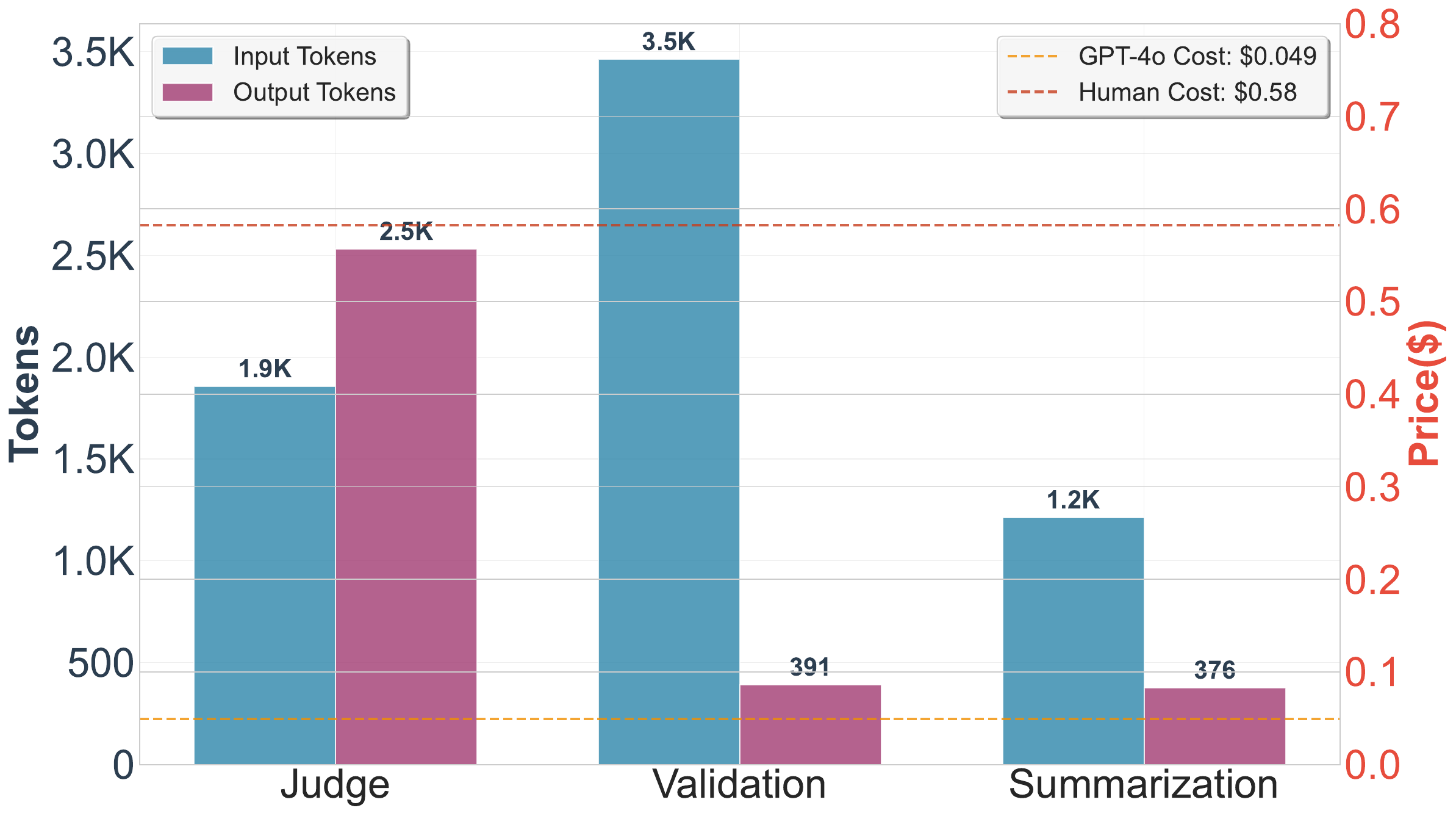}
    \caption{Per-sample token consumption analysis during reward tree construction phase on the StrategyQA dataset.}
    \label{fig:token_analysis_chart}
    \vspace{-1em}
\end{figure}

\paragraph{Reward Quantity Analysis.} 
In Table~\ref{tab:ablation_comparison}, we analyze the reward selection for each step across different datasets. The average number of reward criteria per step is approximately 3.7, comprising an average of 1.5 coarse-grained reward criteria and 2.3 fine-grained reward criteria per step. We observe that for the QASC dataset, coarse-grained reward criteria are entirely sufficient to meet the requirements, while more specialized datasets such as ChemistryQA necessitate a greater number of fine-grained reward criteria.

\vspace{-.5em}
\paragraph{Pareto-based Filtering.} 
We further conduct an ablation study on the effectiveness of Pareto-based filtering. As shown in Table~\ref{tab:ablation_comparison}, we establish the w/o Pareto-based filtering setting by randomly pairing the highest-scoring sample with other samples, ensuring the same number of pairs as in the w. Pareto-based filtering configuration. We observe that employing Pareto dominance effectively improves model accuracy, achieving over 4\% performance improvement on the ARC-c task. Pareto dominance consistently ensures that for each reward criterion $\hat{y}^{(t)}_+ \succ \hat{y}^{(t)}_-$, thereby providing the model with more explicit optimization and learning objectives. Additionally, we analyze the proportion of samples used for pair construction after Pareto-based filtering relative to all samples. We observe that this ratio exceeds 0.5 across all datasets, indicating that DG-PRM demonstrates efficient utilization of sampled steps.

\begin{figure}
    \centering
    \includegraphics[width=\linewidth]{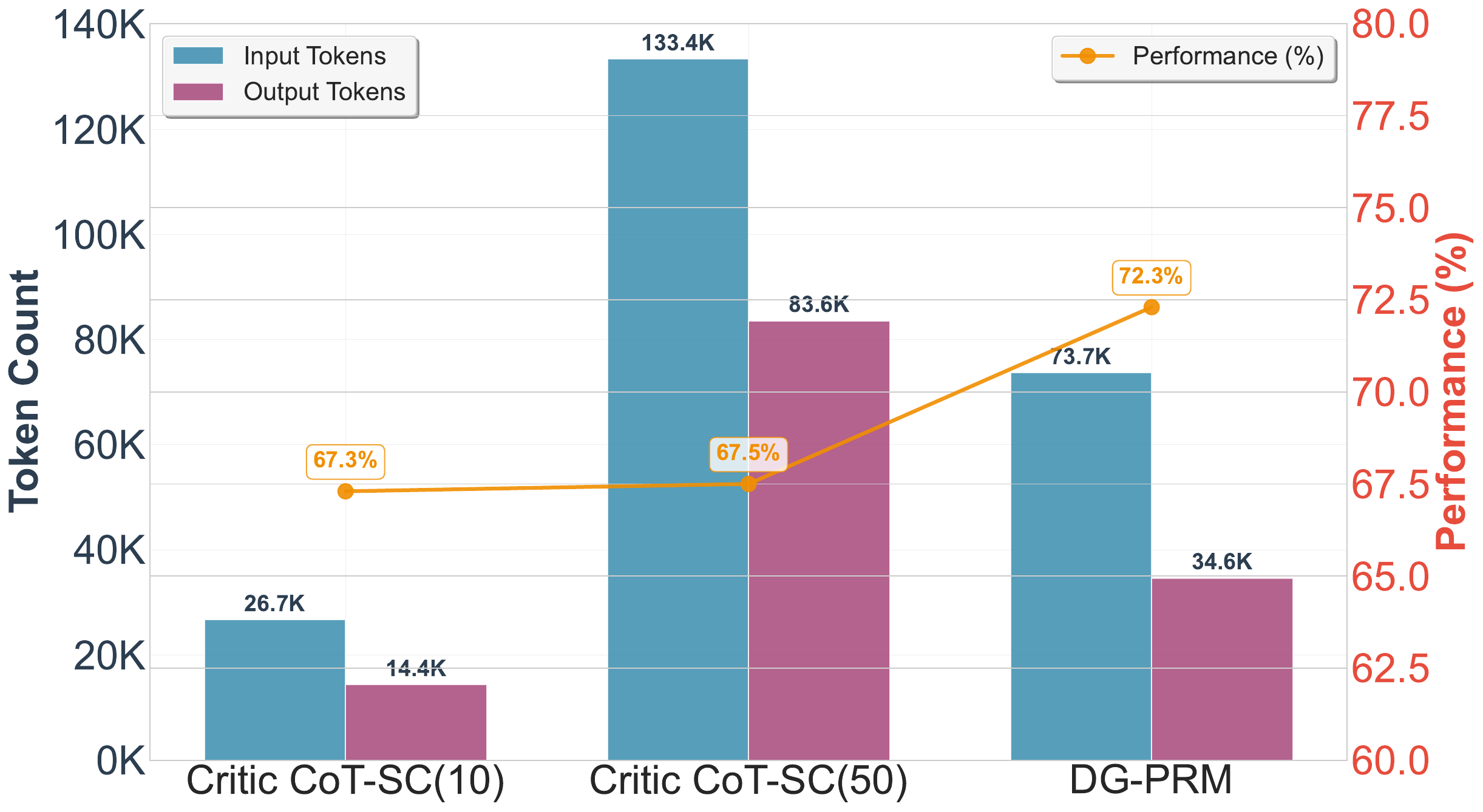}
    \caption{Token consumption and performance comparison of different process reward modeling methods on the PRMBench dataset.}
    \label{fig:critic_methods_analysis_chart}
    \vspace{-1em}
\end{figure}

\vspace{-.5em}
\paragraph{Computational Cost Analysis.} 
In Figure~\ref{fig:token_analysis_chart}, we analyze the costs associated with the reward tree construction phase. The construction costs primarily encompass the judge overhead for generating process reward criteria (Judge), the validation overhead for filtering low-quality criteria (Validation), and the summarization overhead for coarse-grained reward criteria (Summarization). We observe that for a single sample, the input overhead amounts to approximately 7,000 tokens, while the output overhead is around 3,000 tokens. Using the GPT-4o model with official pricing~\footnote{\url{https://openai.com/api/pricing/}}, we find that DG-PRM's cost is substantially lower than manual annotation costs. Furthermore, once the reward tree construction is completed, it can be continuously utilized and updated owing to its generalizability.

We further analyze the overhead during the process reward modeling phase, as illustrated in Figure~\ref{fig:critic_methods_analysis_chart}, where we compare against Critic CoT-SC that employs multiple sampling with majority voting to determine the final output. We observe that simple Self-Consistency~\citep{wang2023selfconsistency} fails to effectively improve the model's step-wise judgment accuracy. Compared to Critic CoT-SC(50), DG-PRM requires only half the computational overhead while achieving a 4.8\% overall performance improvement on PRMBench. This can be attributed to DG-PRM's dynamic reward matching mechanism, which matches relevant and appropriate process reward criteria for each step to compute process reward scores. Compared to Critic CoT, the process reward scores provided are more targeted and facilitate easier traceability of the rationale behind the reward scores.

\section{Conclusion}
\label{sec:conclusion}
In this paper, we first identify that current PRMs are often domain-specific and optimized for specific objectives, limiting their generalizability.
The core challenge lies in the need for detailed, step-by-step supervision, making it difficult to standardize reward signals.
To address this, we propose Dynamic and Generalizable Process Reward Modeling (DG-PRM), 
which utilizes a reward tree to store fine-grained reward signals and dynamically provides step-specific rewards.
We also introduce Pareto dominance estimation to select positive and negative feedback from diverse reward signals. 
Experiments across benchmarks demonstrate that DG-PRM can deliver more accurate process reward signals. Notably, it achieves new state-of-the-art on \textsc{PRMBench}, 
and shows effectiveness in improving model performance across various domains, showcasing its exceptional generalizability by providing effective and precise reward signals.


\section*{Limitations}
\label{sec:limitations}
\paragraph{Expansion to Diverse Domains.} Although we have demonstrated the effectiveness of DG-PRM across several datasets, including \textsc{PRMBench}~\citep{song2025prmbench}, MT-Bench~\citep{zheng2023judging}, and Arena-Hard~\citep{arenahard2024}, there is still significant potential for DG-PRM to expand into more domains. As LLMs are increasingly applied to various scientific fields, DG-PRM can be adapted to design domain-specific reward trees for fields such as drug discovery~\citep{zheng2024largelanguagemodelsdrug}, disease diagnosis~\citep{zhou2024largelanguagemodelsdisease}, and weather forecasting~\citep{wang2024exploringlargelanguagemodels}. We view these applications as exciting avenues for future research.

\paragraph{Incorporating Human External Supervision.} While DG-PRM constructs reward trees using LLMs, and although we have incorporated an automated validator to remove low-quality reward criteria, there remains the possibility of reward criteria that do not align with human expectations. This could lead to potential risks in optimized models. Therefore, it is crucial to introduce appropriate human supervision to refine and enhance the reward tree, and even to design and construct reward trees specific to certain domains.

\paragraph{Adaptation to Advanced Models.} The primary strength of PRM lies in providing reliable and effective process signals for complex reasoning tasks. However, the most advanced models with deep reasoning capabilities, such as OpenAI’s o1~\cite{openai2024o1mini} and o3~\citep{openai2025o3mini}, cannot be adapted due to their closed-source nature. A promising alternative is the DeepSeek R1 model~\citep{guo2025deepseek}. However, due to resource constraints, we have only utilized the distilled version of the model, rather than the full 671B R1 model. In the future, we plan to extend DG-PRM’s adaptability to more advanced models, further enhancing the effectiveness of generalized process supervision.

\section*{Ethics Statement}
\label{sec:ethics_statement}
\paragraph{Compliance with Dataset Licenses.} 
We strictly adhere to the licenses of the datasets used in our experiments. All datasets are in English, and we take care to ensure that our usage aligns with the intended use of each dataset. A detailed overview of the license information for each dataset is provided in Table~\ref{tab:dataset_statistic}.

\paragraph{Adherence to Model Usage Terms.} 
Throughout the experimental process, we strictly follow the terms of use for the models. We comply with the usage guidelines set for models, including terms of service and API usage policies for commercial models. For open-source models, we adhere to the licenses and usage constraints outlined.

\paragraph{Data Privacy.} 
Our method constructs reward trees using the output of LLMs without collecting personal information or sensitive data. We have thoroughly reviewed the prompts and data used in the experiments to ensure they do not contain any personally identifiable information or offensive content.

\paragraph{Data Annotation.} 
During the experiments, we invited five annotators with a master's degree or higher to label the process rewards of model outputs and assess DG-PRM outputs. One of the annotators, a PhD from the Chemistry Department, specifically handled ChemistryQA tasks. Compensation was provided to participants based on local hourly wage standards. All annotators were from the broader Pacific Rim region, with a balanced gender ratio, and we ensured that cultural preferences across different regions were accounted for in the evaluations, ensuring the validity of the results. The instructions provided to annotators are shown in Table~\ref{tab:annotators_2} and Table~\ref{tab:annotators_3}.

\paragraph{Use of AI Assistants.} 
In the evaluation process, we employed AI tools to assist with analyzing model outputs. Specifically, we utilized GitHub Copilot to assist in coding. We ensured that the use of AI tools followed submission guidelines and ethical standards.

\paragraph{Environmental Protection.} 
Training and scaling during testing require significant computational power and resources. Efficient and accurate reward signals enable more efficient model training and allow for the elimination of lower-quality paths during inference, promoting the sustainable development of AI. This approach helps reduce carbon emissions and supports environmental protection.

\section*{Acknowledgment}
\label{sec:acknowledge}
This work was supported by the National Natural Science Foundation of China (No. U24B20181). The computations in this research were performed using the CFFF platform of Fudan University.

\bibliography{custom}

\appendix
\section{Dataset Details}
\label{app:dataset_details}
\begin{table*}[t]
\centering
\footnotesize
\begin{tabular}{l|c|c|c|c|c}
\toprule
\textsc{Dataset} & \textsc{Task} & \textsc{Answer Format} & \textsc{\# Train.} & \textsc{\# Test.} & \textsc{License} \\
\midrule
\href{https://prmbench.github.io/}{\textsc{PRMBench}}~\citep{song2025prmbench} & PRM & Incorrect Position & - & 6,216 & Apache license 2.0 \\
\href{https://huggingface.co/spaces/lmsys/mt-bench/}{MT-Bench}~\citep{zheng2023judging} & General & Rating & - & 80 & Apache license 2.0 \\
\href{https://lmsys.org/blog/2024-04-19-arena-hard/}{Arena-Hard}~\citep{arenahard2024} & General & Rating & - & 500 & Apache license 2.0 \\
\href{https://github.com/allenai/qasc/}{QASC}~\citep{khot2020qasc} & Science & Multi-choice & 8,134 & 926 & Apache license 2.0 \\
\href{https://leaderboard.allenai.org/qasc/submissions/public}{ChemistryQA}~\citep{wei2021chemistryqa} & Science & Text & 2,721 & 392 & CC0-1.0 \\
\href{https://github.com/eladsegal/strategyqa}{StrategyQA}~\citep{geva2021strategyqa} & Commonsense & T/F & 2,061 & 229 &  MIT license \\
\href{https://huggingface.co/datasets/allenai/ai2_arc}{ARC-c}~\citep{clark2018think}  & Commonsense & Multi-choice & 1,119 & 299 &  CC BY-SA 4.0 \\

\bottomrule
\end{tabular}
\caption{Detailed description of the datasets used in the experiments. \textsc{\# Train.} represents the number of training samples, and \textsc{\# Test.} indicates the number of samples used for evaluation.}

\label{tab:dataset_statistic}
\vspace{-1em}
\end{table*}

\begin{table}[t]
\centering
\resizebox{\linewidth}{!}{
    \begin{tabular}{lcc}
    \toprule
    \textbf{Full Name} & \textbf{Evaluation Subject} & \textbf{Number} \\
    \midrule
    Non-Redundancy (NR.) & Simplicity & 758 \\
    Non-Circular Logic (NCL.) & Simplicity & 758 \\
    Empirical Soundness (ES.) & Soundness & 757 \\
    Step Consistency (SC.) & Soundness & 758 \\
    Domain Consistency (DC.) & Soundness & 757 \\
    Confidence Invariance (CI.) & Soundness & 757 \\
    Prerequisite Sensitivity (PS.) & Sensitivity & 756 \\
    Deception Resistance (DR.) & Sensitivity & 750 \\
    Multi-Solution Consistency (MS.) & Sensitivity & 165 \\
    \bottomrule
    \end{tabular}
}
\caption{Statistics of classes in MT-Bench.}
\vspace{-1em}
\label{tab:prmbench_details}
\end{table}
In our experiments, we selected seven datasets encompassing a wide variety of task types that require intricate and complex reasoning by the models. Detailed information on the sample sizes, sources, and licenses of these datasets is provided in Table~\ref{tab:main_results}.

\begin{itemize} 
\item \textbf{\textsc{PRMBench}} is evaluated using the PRM-Score, which is the average of the F1 Score and Negative F1 Score, with an emphasis on the steps where errors occur. Table~\ref{tab:prmbench_details} provides a detailed list of the abbreviations for each category, their corresponding full names, the evaluation objectives, and the number of instances within each class. An example is shown in Table~\ref{tab:prm_samples}.

\item \textbf{MT-Bench}~\citep{zheng2023judging} is evaluated on a 1-10 scale. To ensure comprehensive evaluation and realistic scenarios, we employed a multi-turn setup with evaluation scores output by GPT-4o~\citep{openai2024gpt4o}. We also report the win rates against the GPT-4o model. Table~\ref{tab:mtbench_details} presents the various types and their corresponding quantities in MT-Bench, and Table~\ref{tab:mtbench_samples} illustrates an example from MT-Bench.

\item \textbf{Arena-Hard}~\citep{arenahard2024} uses win rates against the GPT-4-0314 output as an evaluation metric. We used the official output results from \href{https://github.com/lmarena/arena-hard-auto/blob/main/data/arena-hard-v0.1/model_judgment/gpt-4-1106-preview/gpt-4-0613.jsonl}{Arena-Hard’s official repository}, utilizing GPT-4o~\citep{openai2024gpt4o} as the judge model. Arena-Hard consists of 250 diverse scenarios. An example is shown in Table~\ref{tab:arean_samples}.

\item \textbf{QASC}~\citep{khot2020qasc} is a multiple-choice science dataset that provides multiple fact explanations for each answer, aiding LLMs in evaluating the causes of errors from the perspective of facts. A simple example is shown in Table~\ref{tab:QASC_samples}.

\item \textbf{ChemistryQA}~\citep{wei2021chemistryqa} is a chemistry dataset collected by \href{https://socratic.org/chemistry}{Socratic}, covering over 200 topics. It provides the necessary knowledge, conditions, and detailed solution steps. A sample is shown in Table~\ref{tab:ChemistryQA_samples}.

\item \textbf{StrategyQA}~\citep{geva2021strategyqa} challenges models to solve implicit reasoning problems strategically. It provides question breakdowns and corresponding facts, which help identify the points of failure. A simple example is shown in Table~\ref{tab:strategyqa_samples}.

\item \textbf{ARC-c}~\citep{clark2018think} is a challenge subset of the ARC dataset, assessing a model's fundamental reasoning abilities across various fact types, such as Basic Facts \& Properties, and Processes \& Causal.

\end{itemize}

Since \textsc{PRMBench}, MT-Bench, and Arena-Hard do not provide training datasets suitable for constructing reward trees, we leverage the MATH dataset~\citep{hendrycksmath2021} to train \textsc{PRMBench}’s reward tree. For MT-Bench and Arena-Hard, we select 5,000 samples from the LMSYS-Human-Preference-55k~\citep{chiang2024chatbot} dataset to build the reward trees.

\begin{table}[t]
\centering
\resizebox{\linewidth}{!}{
    \begin{tabular}{lcc}
    \toprule
    \textbf{Task Category} & \textbf{Evaluation Focus} & \textbf{Number of Samples} \\
    \midrule
    Writing & Text Generation & 10 \\
    Roleplay & Interaction & 10 \\
    Reasoning & Logical Analysis & 10 \\
    Math & Mathematical Problem Solving & 10 \\
    Coding & Programming & 10 \\
    Extraction & Information Retrieval & 10 \\
    STEM & Scientific Knowledge & 10 \\
    Humanities & Cultural Understanding & 10 \\
    \bottomrule
    \end{tabular}
}
\caption{Statistics of categories in MT-Bench.}
\label{tab:mtbench_details}
\end{table}

\section{Experiment Details}
\label{app:experiment_details}

\subsection{Implementation Details}
\label{app:implement_details}
In the process, we segment steps using newline characters or explicit labels such as ``Step1'' and ``Step2.'' For each step, we construct positive and negative label pairs based on the reference answers \( y \) or evaluation criteria \( \mathcal{C} \) provided in the dataset, such as MT-Bench~\citep{zheng2023judging}. We use the corresponding judge prompt and the reference answer \( y \) from the dataset to build these pairs. 

We utilize GPT-4o to construct the judge model \( \mathcal{J} \), generating appropriate reward criteria for erroneous steps. The BAAI/bge-en-icl~\citep{li2024makingtextembeddersfewshot} embedding model is employed to obtain embedding vectors \( v \), where the vector space has a dimension \( d = 4096 \). Additionally, in our experimental analysis, we incorporate the text-embedding-3-large~\citep{openai2024embedding} and nvidia/NV-Embed-v2~\citep{lee2024nv} models. 

The BIRCH algorithm~\citep{zhang1997birch} is used for hierarchical clustering, with the Birch implementation from \texttt{sklearn.cluster}. In the process of merging reward criteria, we retain criteria with longer lengths. Following the clustering, GPT-4o is used to summarize the upper-level coarse-grained reward criteria \( r^{\text{parent}} \). This step proves crucial for subsequent reward allocation, as using higher-quality models for summarization yields more accurate reward signals.

The same model used to generate the coarse-grained reward criteria is employed for reward allocation, determining whether fine-grained reward criteria \( r^{\text{child}} \) should be provided. To ensure the accuracy of evaluation, we sample multiple times to confirm the appropriateness of selected reward criteria. In our experiments, we set the sampling frequency to 5, retaining coarse-grained reward criteria that appear more than 3 times. Typically, the model's judgments are consistent.

During scoring, previously generated steps, evaluation information, and scores are concatenated at the front for reference. By default, we use GPT-4o as the scorer \( \mathcal{S} \). We also analyze the evaluation results using the model's own evaluation and the o3-mini evaluation results, as detailed in \ref{app:judge_model}.

If corresponding positive or negative feedback cannot be found, additional sampling is performed until the maximum number of attempts is reached. 
Table~\ref{tab:DG-PRM} provides an example with selected fine-grained reward criteria placed in the prompt. In practice, only coarse-grained reward criteria can be included due to context limitations, so we carefully select fine-grained reward criteria based on matching. In Table~\ref{tab:coarse_criteria}, Table~\ref{tab:fine_criteria_1}, and Table~\ref{tab:fine_criteria_2}, we present several examples of coarse-grained and fine-grained process reward criteria.

For training, we use 8 interconnected H200 GPUs with a batch size of 32, a learning rate of 5e-6, and a DPO-beta of 0.1. Other settings are based on the default parameters of \texttt{HuggingFace's DPOTrainer}. Due to the temporary unavailability of the DeepSeek-R1 API, we deploy the full 671B version using \texttt{ollama} on an 8 interconnected A100 setup with INT4 precision. The same approach is used for the other Distill models.

During generation, we adjust the temperature parameter \( \gamma \) according to the task and model. Specifically, we find that for R1-Distill-Qwen-1.5B and R1-Distill-Qwen-7B, higher temperatures lead to very long sequences and many repeated solutions, as observed by \citet{chen2025think23overthinkingo1like} in the QwQ~\citep{qwq-32b-preview} model. Thus, we set \( \gamma \in [0.5, 0.6] \) and filter out sequences exceeding 4096 tokens. For R1-Distill-Qwen-14B and R1-Distill-Qwen-32B, we find that \( \gamma \in [0.7, 0.8] \) yields more satisfactory results on MT-Bench and Arean-Hard, while a lower temperature works better for QASC, StrategyQA, and ARC-c tasks, so we set \( \gamma \in [0.6, 0.7] \). For the MT-Bench and Arean-Hard datasets, we compute the average score of five output results. For other datasets, we calculate the final results based on the corresponding metrics.

\subsection{Automated Validator}
\label{app:automated_validator}

During the construction of the reward tree \(\mathcal{T}\), an automated validator plays a crucial role in ensuring the high quality of the reward criteria. This validator can be applied to the reward tree construction for any model, filtering out low-quality reward criteria. We use GPT-4o~\citep{openai2024gpt4o} as the automated validator, which evaluates the reward criteria through prompts and provides one of three assessment results: \texttt{Good}, \texttt{Ordinary} and \texttt{Bad}. Detailed criteria for these evaluations are outlined, and Table~\ref{tab:automated_validator} provides a general example.

\begin{figure}[t]
  \centering
  \includegraphics[width=\linewidth]{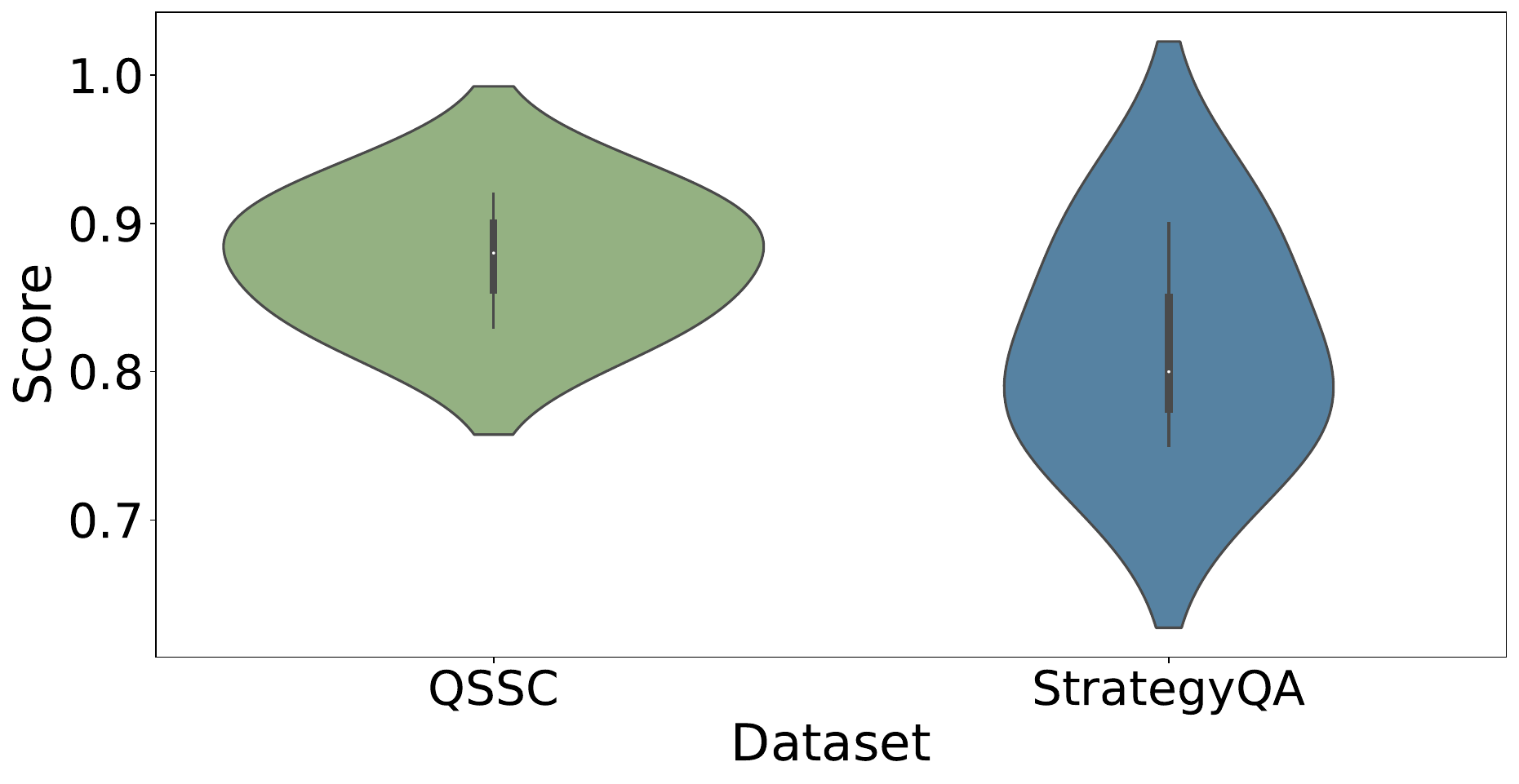}
    \caption{Consistency between automated validator and human evaluation}
  \label{fig:automated_validator}
  \vspace{-1em}
\end{figure}

To assess the consistency between the automated validator and human evaluations, we randomly selected 100 outputs from the automated validator and compared them to the results of human evaluations. We calculated the proportion of matches between the automated outputs and the labels assigned by three human evaluators, with the results shown in Figure~\ref{fig:automated_validator}. Our analysis reveals that, on both the QASC and StrategyQA datasets, the automated validator exhibits high consistency with human evaluations.

\begin{figure*}[t]
  \centering
  \begin{subfigure}{0.32\linewidth}
    \includegraphics[width=\linewidth]{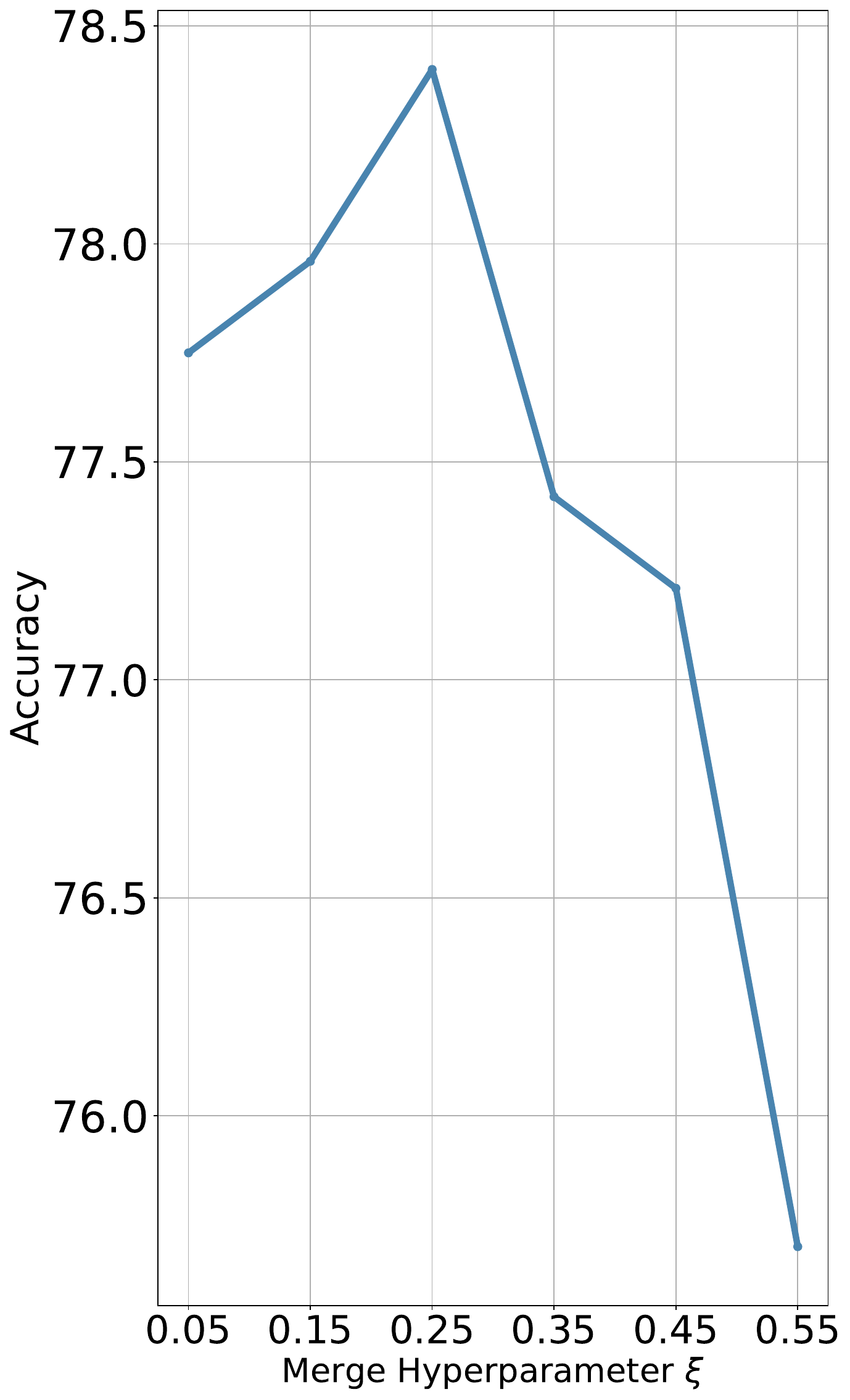}
    \caption{Merge Hyperparameter \(\xi\)}
    \label{fig:merge_hyperparameter}
  \end{subfigure}
  \hfill 
  \begin{subfigure}{0.32\linewidth}
    \includegraphics[width=\linewidth]{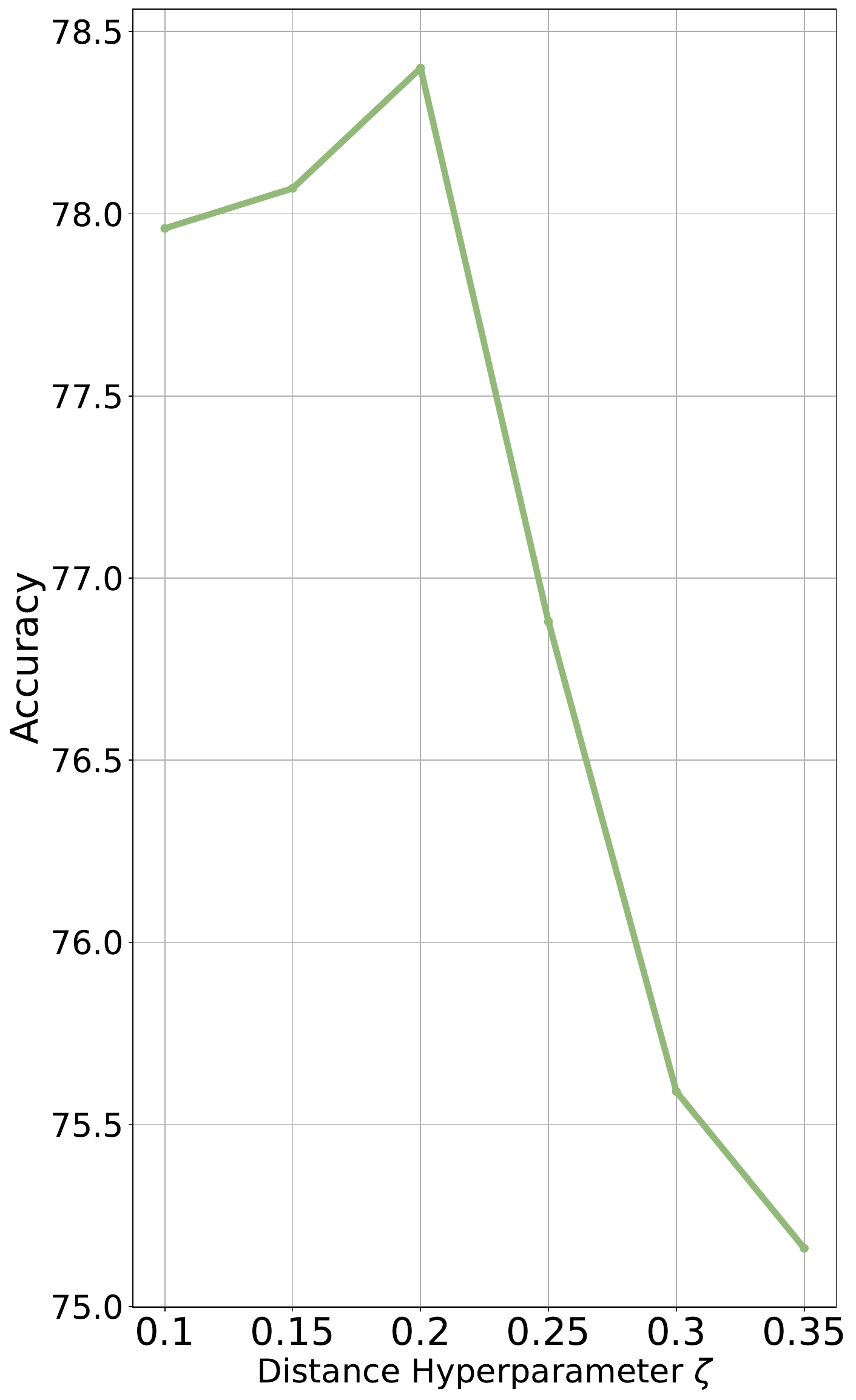}
    \caption{Distance Hyperparameter \(\zeta\)}
    \label{fig:distance_hyperparameter}
  \end{subfigure}
  \hfill 
  \begin{subfigure}{0.32\linewidth}
    \includegraphics[width=\linewidth]{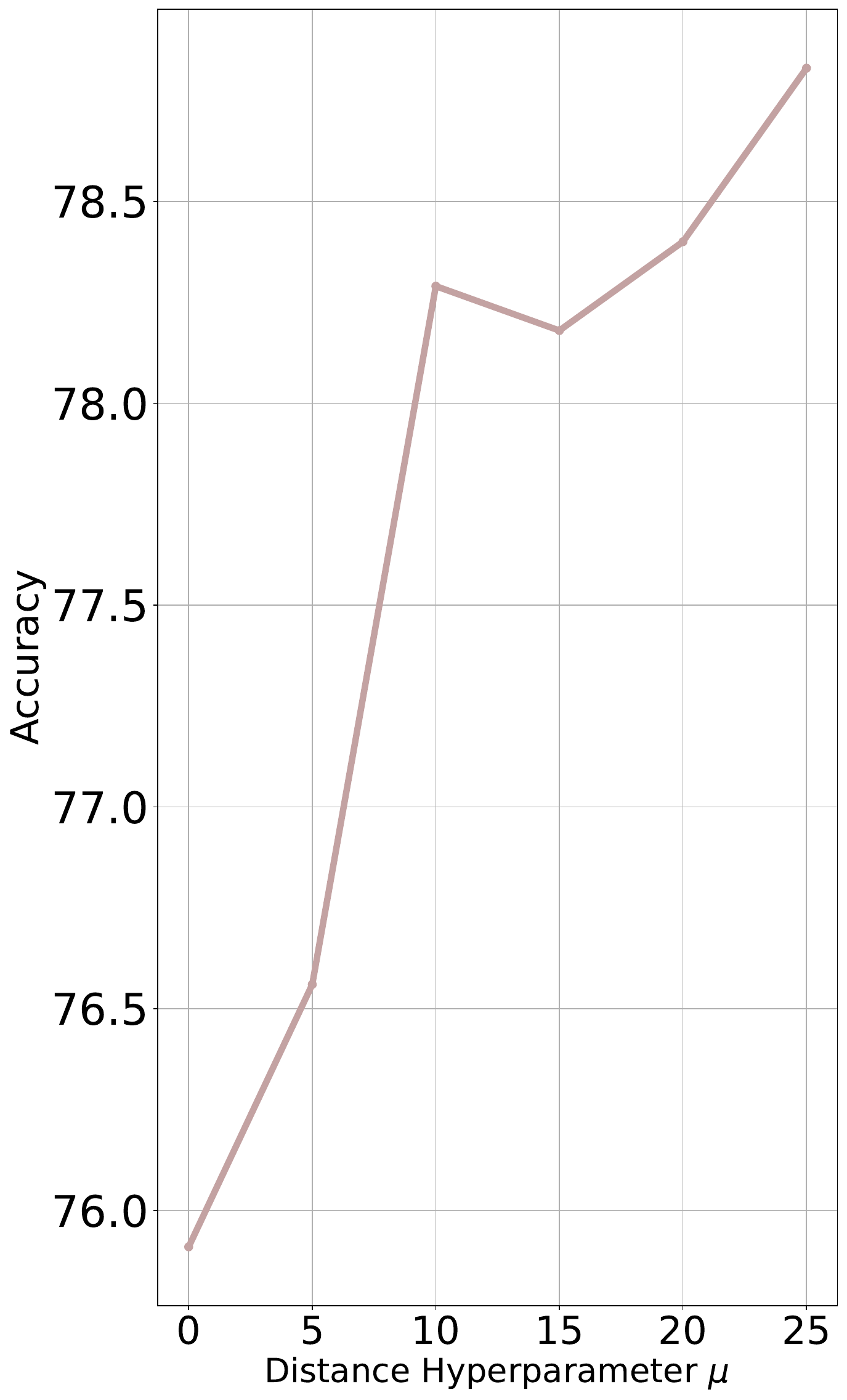}
    \caption{Step Hyperparameter \(\mu\)}
    \label{fig:step_hyperparameter}
  \end{subfigure}
    \caption{Ablation analysis of the hyperparameters on the QASC dataset, using R1-Distill-Qwen-7B as the backbone.}
\vspace{-1em}
\label{fig:hyperparameter_analysis}
\end{figure*}

\subsection{Baseline Implementation}
\label{app:baseline_implement}
For the ORM, we follow the approach of \citet{liu2024skywork} and train the ORM based on the Qwen-2.5-7B model~\citep{yang2024qwen2}. We employ AdamW as the optimizer, with a batch size of 16 and a learning rate of 2e-6. In the process of building Critic Models, we prompt the LLM to perform step-by-step evaluation. Specifically, the Critic Direct directly outputs whether each step is correct, while the Critic CoT provides a detailed analysis of the output steps before giving the final result. Each step is evaluated with \texttt{[[correct]]} or \texttt{[[wrong]]}, and the final judge model extracts any erroneous steps. If all steps are correct, the output is \texttt{[]}. Tables~\ref{tab:critic_direct} and~\ref{tab:critic_cot} show an example of the Critic Direct and Critic CoT.

\section{Hyperparameter Analysis}
\label{app:hyperparameter_analysis}
We conducted an ablation study on the hyperparameters used in our experiments on the QASC dataset, as shown in Figure~\ref{fig:hyperparameter_analysis}.

\paragraph{Merge Hyperparameter \(\xi\)} The hyperparameter \(\xi\) controls the merging of similar reward criteria. As shown in Figure~\ref{fig:merge_hyperparameter}, a larger value of \(\xi\) leads to the merging of related criteria, and in an extreme case, all \(r^{\text{child}}\) under \(r^{\text{parent}}\) are merged, resulting in a coarse-grained process reward. This has a significant impact on performance. On the other hand, a smaller value of \(\xi\) leads to a large number of redundant criteria in the reward tree \(\mathcal{T}\), which increases noise and causes performance degradation. It also results in redundant computations in \(\mathcal{S}\), leading to unnecessary overhead. Therefore, we set \(\xi = 0.25\) as a suitable threshold.

\vspace{-.5em}
\paragraph{Distance Hyperparameter \(\zeta\)} The hyperparameter \(\zeta\) controls the selection of fine-grained process reward criteria, as illustrated in Figure~\ref{fig:distance_hyperparameter}. A larger value of \(\zeta\) results in the inclusion of numerous irrelevant reward criteria, leading to a decrease in overall performance. This occurs because an excessive number of criteria makes it difficult for DG-PRM to select the corresponding positive and negative samples. Consequently, it is essential to limit the criteria within an optimal range. On the other hand, a smaller value of \(\zeta\) may fail to match the appropriate fine-grained process reward criteria, similarly impairing performance.

\vspace{-.5em}
\paragraph{Step Hyperparameter \(\mu\)} The hyperparameter \(\mu\) controls the number of steps within the reward criteria selection \(\mathcal{R}\) and scoring \(\mathcal{S}\) that can be referenced. We observe that as the number of steps increases, performance continues to improve. Therefore, providing a larger number of steps is beneficial for the accuracy of DG-PRM selection and scoring. However, considering the constraints of the model's context window, we do not set \(\mu\) above 20. In the ChemistryQA scenario, the reasoning process is more complex, which can exceed the model's context window. Additionally, we observe that performance gains gradually diminish. Therefore, considering the cost overhead, we set \(\mu = 20\).

\section{Analysis and Discussion}
\label{app:analysis_and_discussion}
\subsection{Judge Model}
\label{app:judge_model}
\begin{figure}[t]
  \centering
  \includegraphics[width=0.9\linewidth]{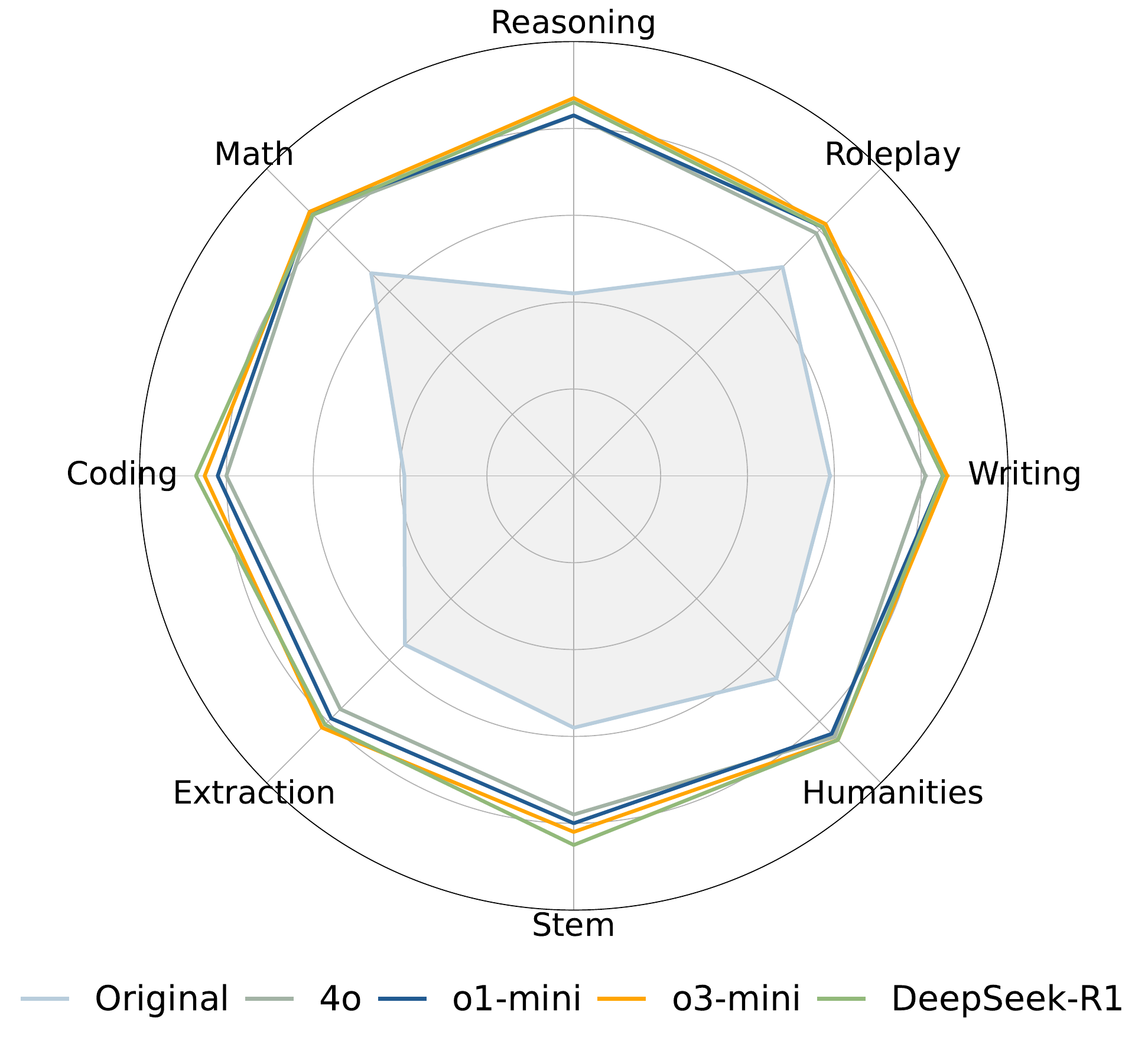}
    \vspace{-.1em}
    \caption{Impact of the judge model on ratings in the MT-Bench dataset, using R1-Distill-Qwen-7B as the backbone.}
    \vspace{-.5em}
  \label{fig:judge_model}
\end{figure}

In Figure~\ref{fig:judge_model}, we analyze the impact of different models as \(\mathcal{R}\) and \(\mathcal{S}\) on performance using the MT-Bench dataset. We utilize radar charts to display the variation in ratings across each category. 
Our observations reveal that employing a better judge model can significantly enhance performance in writing, reasoning, and coding tasks, indicating that DG-PRM can continuously benefit from a superior judge model. Furthermore, we find that DeepSeek-R1 exhibits performance comparable to o3-mini, even outperforming it in coding and STEM tasks. This demonstrates that DG-PRM is also applicable to advanced open-source models.

\subsection{Implicit PRM}
\label{app:implicit_prm}
\begin{figure}[t]
  \centering
  \includegraphics[width=\linewidth]{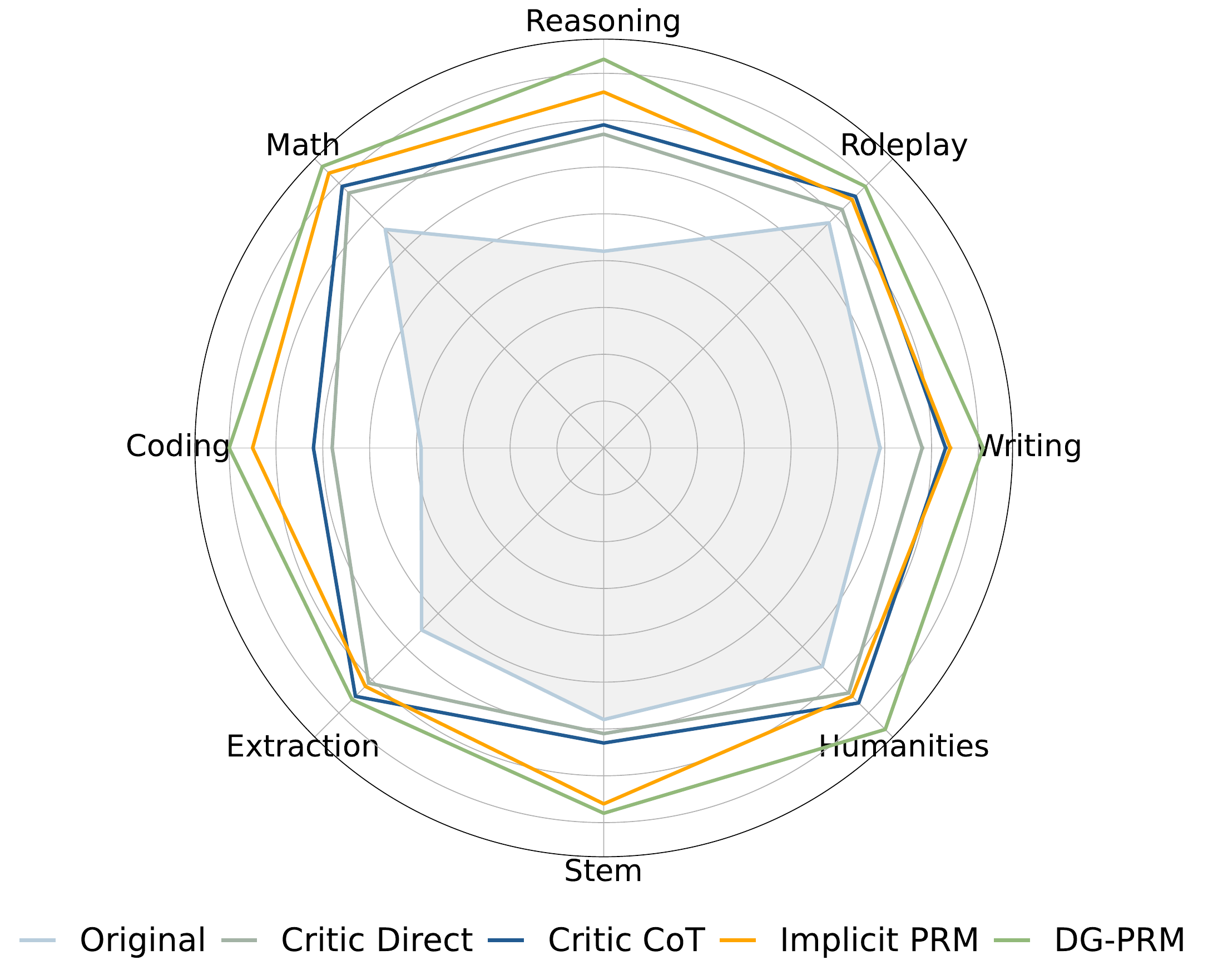}
    \caption{Performance comparison of different PRM methods on the MT-Bench dataset, using R1-Distill-Qwen-7B as the backbone.}
    \vspace{-.5em}
  \label{fig:implicit_prm}
\end{figure}

In Figure~\ref{fig:implicit_prm}, we compare the performance of different methods on the MT-Bench dataset. We include Implicit PRM~\citep{rafailov2024from}, which is trained using the LMSYS-Human-Preference-55k~\citep{chiang2024chatbot} dataset based on the official implementation. We regenerate the responses using R1-Distill-Qwen-7B and construct the chosen and rejected pairs based on the scoring from GPT-4o. We observe that Implicit PRM effectively improves performance, particularly in reasoning and coding, suggesting that Implicit PRM can effectively model tasks with well-defined objectives. However, DG-PRM demonstrates more substantial improvements, offering a more comprehensive enhancement of model performance, such as its exceptional results in writing and humanities tasks. Furthermore, DG-PRM is more interpretable, providing a clear explanation of the advantages of positive samples over negative samples, making the optimization objectives easier to understand.

\subsection{Embedding Model}
\label{app:embedding_model}
\begin{table}[t]
\centering
    \begin{tabular}{lcc}
    \toprule
    \textbf{Embedding Model} & \textbf{QASC} & \textbf{ChemistryQA} \\
    \midrule
    BAAI/bge-en-icl & 78.40 & 87.50 \\
    text-embedding-3-large & \textbf{79.04} & \textbf{88.01} \\
    nvidia/NV-Embed-v2 & 78.29 & 87.24 \\
    \bottomrule
    \end{tabular}
\caption{Analysis of embedding models.}
\vspace{-1em}
\label{tab:embedding_model}
\end{table}
Table~\ref{tab:embedding_model} analyzes the impact of different embedding models on performance. The embedding model plays a crucial role in the construction of the reward tree and the selection of fine-grained process reward criteria. We find that DG-PRM demonstrates robustness in terms of embedding model selection, achieving satisfactory performance even with open-source models. Therefore, we use the BAAI/bge-en-icl model as the embedding function \(\mathcal{V}\) in our experiments.


\subsection{Hierarchical Clustering}
\label{app:hierarchical_clustering}
\begin{table}[t]
\centering
    \begin{tabular}{lcc}
    \toprule
    \textbf{Embedding Model} & \textbf{QASC} & \textbf{ChemistryQA} \\
    \midrule
    BRICH & 78.40 & 87.50 \\
    Agglomerative & 78.83 & 86.73 \\
    Divisive & \textbf{79.48} & \textbf{88.26} \\
    \bottomrule
    \end{tabular}
\caption{Analysis of hierarchical clustering algorithm.}
\vspace{-1em}
\label{tab:hierarchical_clustering}
\end{table}
In Table~\ref{tab:hierarchical_clustering}, we examine the effect of different hierarchical clustering methods on performance. We selected two approaches: agglomerative and divisive clustering, using Ward's method~\citep{ward1963hierarchical} to define cluster distances. Our findings reveal that divisive clustering yields better performance. However, given the high computational cost of divisive clustering, the incremental updates offered by the BIRCH algorithm~\citep{zhang1997birch} significantly reduce overhead, making the addition and removal of reward criteria more efficient and convenient. 

Specifically, we observe that the BIRCH algorithm completes hierarchical clustering within 10.2 seconds on PRMBench and 19.53 seconds on ChemistryQA, demonstrating its ability to perform clustering efficiently within 20 seconds across different datasets. Therefore, we selected the BIRCH algorithm as the clustering method to obtain the reward tree.

\begin{figure}[t]
  \centering
  \includegraphics[width=\linewidth]{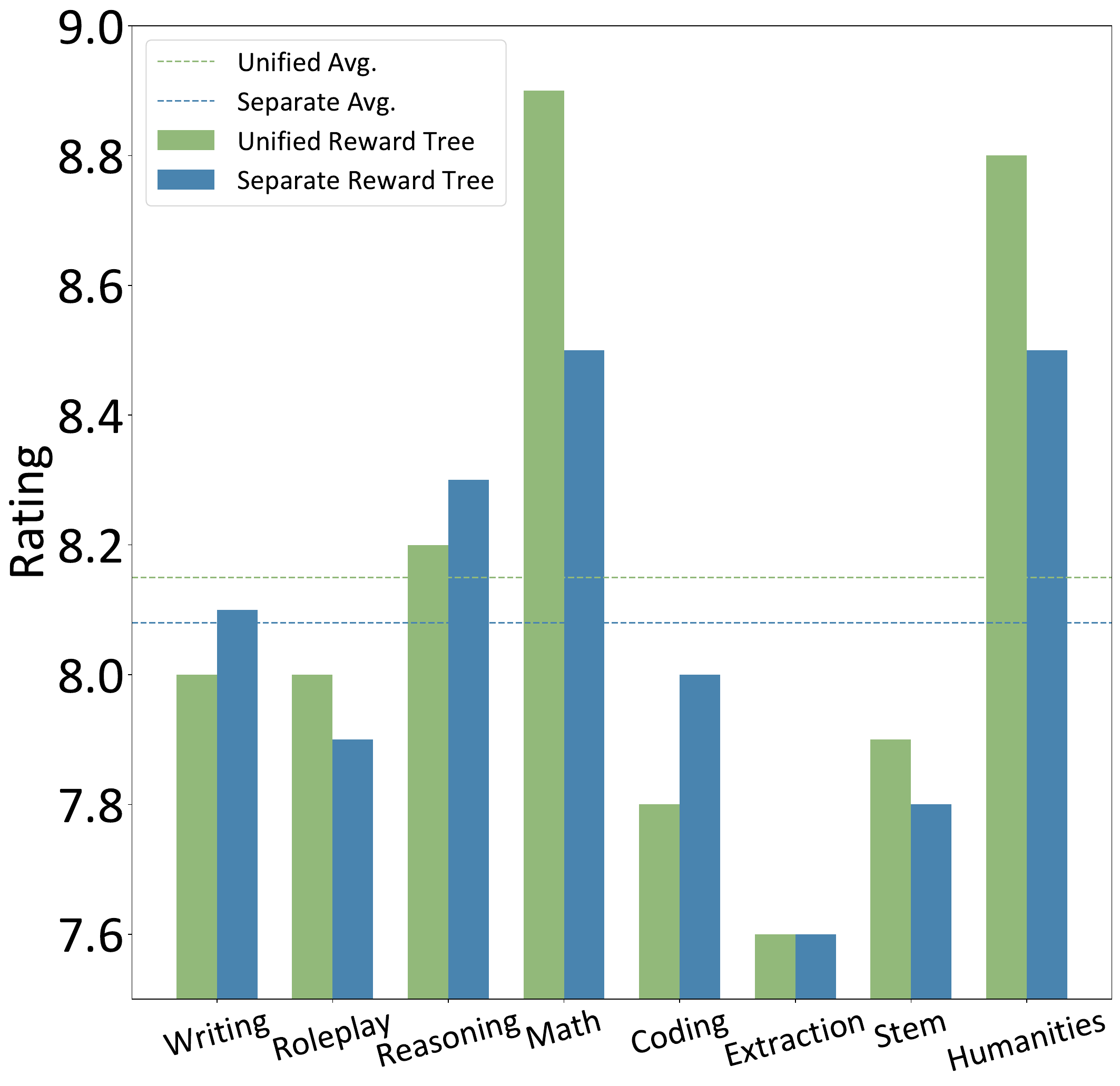}
    \caption{Comparison of ratings on MT-Bench using a unified reward tree versus separate reward trees.
}
  \label{fig:unified_separate_reward_tree}
  \vspace{-1em}
\end{figure}

\subsection{General Reward Tree \(\mathcal{T}\)}
\label{app:general_reward_tree}
To further evaluate the generalizability of DG-PRM, we construct a unified reward tree encompassing a rich set of criteria from the MATH, QASC, Chemistry, StrategyQA, and ARC-c training datasets. Figure~\ref{fig:unified_separate_reward_tree} presents the experimental results using this unified reward tree on MT-Bench. Compared to a reward tree constructed separately for MT-Bench, the unified reward tree demonstrates better performance on humanities and math tasks. We observe that the model incorporates a broader range of factual evaluations during scoring. These criteria likely stem from the diverse mathematical perspectives in MATH and fact-related assessments in StrategyQA. The dashed line represents the average performance of the unified and separate reward trees. Our findings show that the unified reward tree achieves a higher overall score, highlighting the exceptional scalability of DG-PRM.

\subsection{Human Consensus}
\label{app:human_consensus}
\begin{figure}[t]
  \centering
  \includegraphics[width=\linewidth]{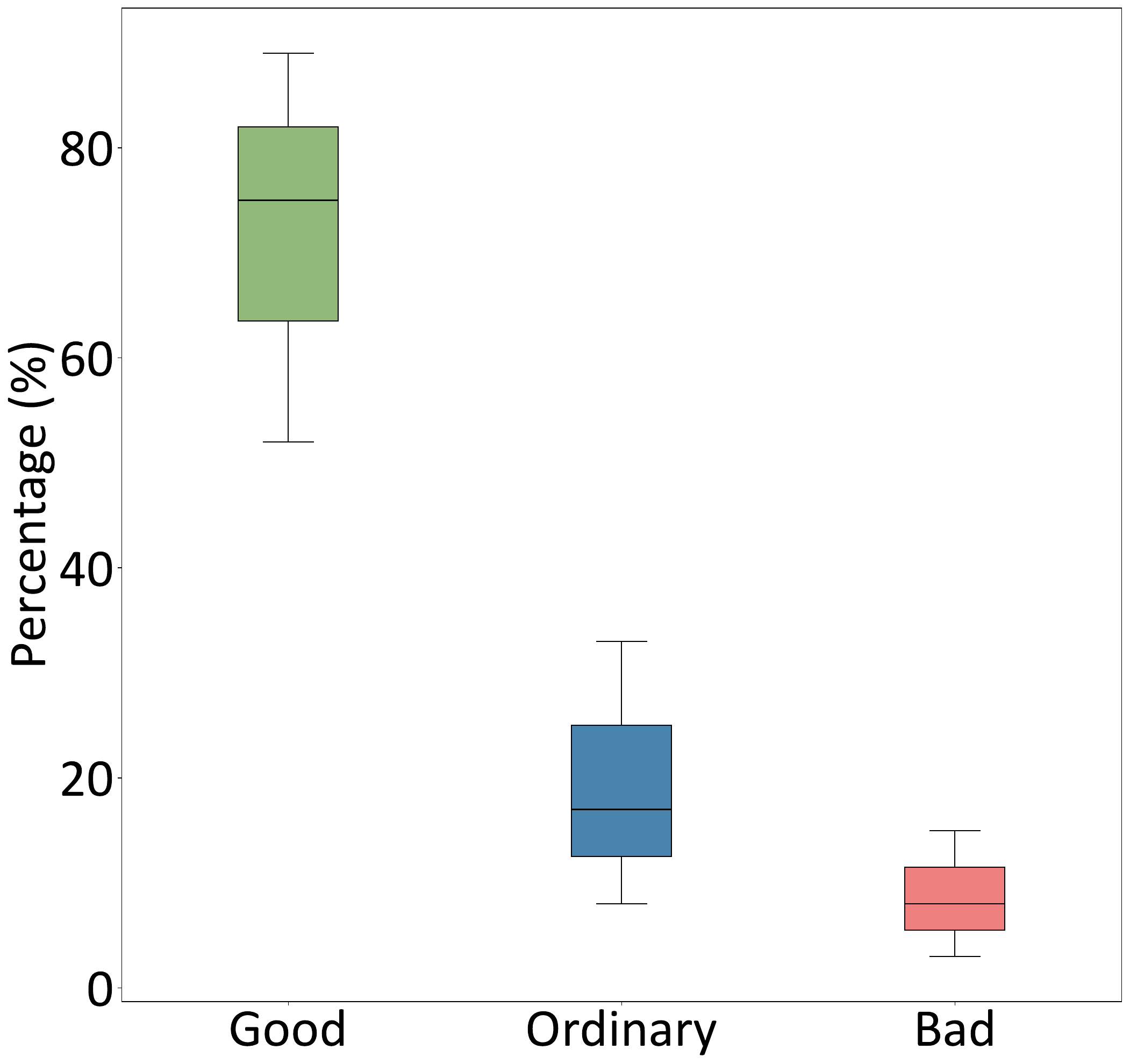}
    \caption{Human evaluation of DG-PRM output, including the selection of reward criteria and scores.
}
  \label{fig:human_consensus}
\end{figure}

To further analyze the rationality of DG-PRM's reward criteria allocation and output reward scores, we randomly select 100 samples, including the problem, answer, assigned criteria, and final reward score. These are independently evaluated by three assessors, who judge the appropriateness of the allocated criteria and output score, labeling each sample as \texttt{Good}, \texttt{Ordinary}, or \texttt{Bad}. The results are shown in Figure~\ref{fig:human_consensus}. We observe that the median proportion of \texttt{Good} labels is close to 80\%, while the proportion of \texttt{Bad} labels is under 10\%, indicating that the majority of assessors find DG-PRM's reward allocation and output scores to be reasonable, demonstrating the rationality of DG-PRM's reward criteria construction and score output.

\subsection{Optimization Algorithm}
\label{app:rl}
\begin{table}[t]
\centering
\resizebox{\linewidth}{!}{
    \begin{tabular}{lcc}
    \toprule
    \textbf{Optimization Algorithm} & \textbf{Single-Turn} & \textbf{Multi-Turn} \\
    \midrule
    Baseline & 6.01 & 5.66 \\
    DPO~\citep{rafailov2023direct} & \textbf{8.67} & \textbf{8.09} \\
    CPO~\citep{xu2024cpo} & 8.58 & 7.87 \\
    SimPO~\citep{meng2024simpo} & 8.62 & 7.95 \\
    \bottomrule
    \end{tabular}
}
\caption{Performance analysis of different optimization algorithms on MT-Bench.}
\vspace{-1em}
\label{tab:rl}
\end{table}
In Table~\ref{tab:rl}, we analyze the performance of DG-PRM using various optimization algorithms. Experimental results on MT-Bench show that the process rewards constructed by DG-PRM are effective across multiple optimization algorithms. Significant performance improvements are observed in both Single-Turn and Multi-Turn scenarios compared to the baseline. Among the algorithms tested, DPO achieves the most notable performance gain, thus we utilize DPO to optimize our policy model in the experiments.

\section*{Visions of Tomorrow}

In recent years, we have witnessed a rapid expansion of model capabilities, with AI increasingly integrating into human scientific research~\citep{openai2023introducing}. This necessitates the development of a generally applicable Process Reward Model (PRM). In the short term, PRM will focus on aligning with the preferences of human scientists and our understanding of the objective world, preventing erroneous search paths and guiding the model toward more meaningful exploration. As model capabilities continue to evolve, PRM will steer the model towards human-desired goals, such as environmental sustainability and disease treatment, rather than facilitating the creation of harmful products in explorations. We believe that a generalizable PRM will play a pivotal role in advancing AI in a safer and more reliable direction.

\begin{table*}
    \centering
    \small
    \begin{tabular}{p{0.95\linewidth}}
    \toprule
        \textbf{Original Process}\\
        \textbf{Question}: If $x^2 - x - 1 = 0$, what is the value of $x^3 - 2x + 1$?\\
        
        \textbf{Step 1}: Let's start with the first equation and see if we can solve for x.\\
        
        \textbf{Step 2}: We can use the quadratic formula to find x.\\
        
        \textbf{Step 3}: The quadratic formula is $x = \frac{-b \pm \sqrt{b^2 - 4ac}}{2a}$.\\
        
        \textbf{Step 4}: Substituting $a = 1$, $b = -1$, and $c = -1$, we get $x = \frac{1 \pm \sqrt{5}}{2}$.\\
        
        \textbf{Step 5}: Since $x^2 - x - 1 = 0$, we replace $x^2$ with $x + 1$ in $x^3 - 2x + 1$.\\
        
        \textbf{Step 6}: Substituting, $x^3 - 2x + 1 = x(x + 1) - 2x + 1$.\\
        
        \textbf{Step 7}: Simplify: $x^3 - 2x + 1 = (x^2 + x) - 2x + 1$.\\
        
        \textbf{Step 8}: Substitute $x^2 = x + 1$: $x^3 - 2x + 1 = (x + 1 + x) - 2x + 1$.\\
        
        \textbf{Step 9}: Simplify: $x^3 - 2x + 1 = 2$.\\
    
        \midrule
        \textbf{Modified Process}\\
        
        \textbf{Question}: If $x^2 - x - 1 = 0$, what is the value of $x^3 - 2x + 1$?\\
        
        \textbf{Step 1}: Let's start with the first equation and see if we can solve for x.\\
        
        \textbf{Step 2}: We can use the quadratic formula to find x.\\
        
        \textbf{Step 3}: The quadratic formula is $x = \frac{-b \pm \sqrt{b^2 - 4ac}}{2a}$.\\
        
        \textbf{Step 4}: Substituting $a = 1$, $b = -1$, and $c = -1$, we get $x = \frac{1 \pm \sqrt{5}}{2}$.\\
        
        \textbf{Step 5}: Since $x^2 - x - 1 = 0$, we replace $x^2$ with $x + 1$ in $x^3 - 2x + 1$.\\
        
        \textbf{Step 6}: Substituting, $x^3 - 2x + 1 = x(x + 1) - 2x + 1$.\\
        
        \textcolor[rgb]{0.25, 0.5, 0.75}{\textbf{Step 7}: To simplify $x^3$, use the property of cubic roots: $x^3 = \sqrt[3]{x}$.}\\
        
        \textbf{Step 8}: Substitute $x^3 = \sqrt[3]{x}$ and $x^2 = x + 1$ into $x^3 - 2x + 1$: $x^3 - 2x + 1 = \sqrt[3]{x} - 2x + 1$.\\
        
        \textbf{Step 9}: Simplify further: $x^3 - 2x + 1 = \sqrt[3]{x} - x + 1$.\\
    
        \midrule
        \textbf{Reason}\\
        
        A domain inconsistency was introduced in Step 7 by incorrectly assuming that $x^3 = \sqrt[3]{x}$. This assumption is valid for root-based equations but not in the context of solving a polynomial equation. This error propagates through the rest of the reasoning, leading to a flawed solution.\\
        \bottomrule
    \end{tabular}
    \caption{\textsc{PRMBench} samples.}
    \label{tab:prm_samples}
\end{table*}

\begin{table*}
    \centering
    \small
    \begin{tabular}{p{0.95\linewidth}}
    \toprule
        \textbf{Question}: Climate is generally described in terms of what?\\
        
        \textbf{Answer Choice}: (A) sand (B) occurs over a wide range (C) forests (D) Global warming (E) rapid changes occur (F) local weather conditions (G) measure of motion (H) city life\\
        
        \textbf{Fact1}: Climate is generally described in terms of temperature and moisture.\\

        \textbf{Fact2}: Fire behavior is driven by local weather conditions such as winds, temperature and moisture.\\

        \textbf{answerKey}: F\\
        \bottomrule
    \end{tabular}
    \caption{QASC samples.}
    \label{tab:QASC_samples}
\end{table*}

\begin{table*}
    \centering
    \small
    \begin{tabular}{p{0.95\linewidth}}
    \toprule
        \textbf{Question}: How many moles of phosphorus trichloride would contain 3.35 times $10^24$ molecules of phosphorus trichloride?\\
        
        \textbf{Conditions}: Number $[OF]$ phosphorus trichloride molecules $[=] pu{3.35 × 10^24}$\\
        
        \textbf{TargetVariable}: Mole $[OF]$ phosphorus trichloride $[IN]$ moles\\

        \textbf{Answer}: 5.56 moles\\

        \bottomrule
    \end{tabular}
    \caption{ChemistryQA samples.}
    \label{tab:ChemistryQA_samples}
\end{table*}

\begin{table*}
    \centering
    \small
    \begin{tabular}{p{0.95\linewidth}}
    \toprule
        \textbf{Question}: Did Aristotle use a laptop?\\
        
        \textbf{Answer}: No\\

        \bottomrule
    \end{tabular}
    \caption{StrategyQA samples.}
    \label{tab:strategyqa_samples}
\end{table*}

\begin{table*}
    \centering
    \small
    \begin{tabular}{p{0.95\linewidth}}
    \toprule
        \textbf{Question}: Juan and LaKeisha roll a few objects down a ramp. They want to see which object rolls the farthest. What should they do so they can repeat their investigation?\\
        
        \textbf{Answer Choice}: (A) Put the objects in groups. (B) Change the height of the ramp. (C) Choose different objects to roll. (D) Record the details of the investigation.\\
        
        \textbf{Answer}: D\\

        \bottomrule
    \end{tabular}
    \caption{ARC-c samples.}
    \label{tab:arc_samples}
\end{table*}

\begin{table*}
    \centering
    \small
    \begin{tabular}{p{0.95\linewidth}}
    \toprule
        \textbf{Question 1}: Compose an engaging travel blog post about a recent trip to Hawaii, highlighting cultural experiences and must-see attractions.\\
        
        \textbf{Answer}
        **A Glimpse into the Enchanted Land of Hawaii: A Cultural and Adventurous Journey**\\
        
        Our recent trip to Hawaii was an absolute blast, a journey that left us thoroughly charmed by its beauty, culture, and unique spirit. As we explored this archipelago of islands, each destination offered a taste of its rich heritage and stunning landscapes.\\
        
        One experience that stood out for me was the traditional Luau at Waikiki Beach Park. What made it truly unforgettable was the opportunity to step into a cultural immersion, dancing to the rhythm of hula while sharing stories with locals. It was like being transported back in time to a vibrant Hawaiian village, where music and hospitality were the foundation of life.\\
        
        Our journey began in Honolulu, where we strolled along Waikiki Beach, soaking in the warm sun and vibrant ocean views. This iconic location isn't just a beach; it's a symbol of Hawaii's blend of tradition and modernity, perfect for capturing photos with the iconic "Hawaii Five-O" lighthouse in the background.\\
        
        Next up was Pearl Harbor, where history took center stage. The experience was deeply moving as we witnessed the power of remembrance at this historic site. It reminded us how crucial it is to honor our heroes and preserve their legacy.\\
        
        We then ventured to Hanauma Bay, a destination steeped in nature's beauty. Scattered with fronds and lined by sea grasses, this bay offered a serene escape, allowing us to truly disconnect from the hustle and bustle of city life.\\
        
        Finally, we visited the Ironman Training Center at Waikoloa State Park, challenging ourselves physically and mentally. The rigorous workout left us exhilarated, pushing our limits while fostering camaraderie with fellow participants.\\
        
        Reflecting on this trip, I can say that Hawaii's culture is as vibrant as its landscapes. Every experience was a lesson in appreciation for life's beauty and the richness of diverse traditions. Our time there will forever be cherished, not just for the sights and sounds, but for the profound impact it had on our hearts.\\
        
        Hawaii's magic continues to inspire us, and we can't wait to return. If you're ever in the mood for an adventurous adventure with a touch of culture, Hawaii is the perfect destination!\\
    
        \midrule
        
        \textbf{Question 2}: Rewrite your previous response. Start every sentence with the letter A.\\
        
        \textbf{Answer}: 
        
        **A Clever Rewrite: A Glimpse into the Enchanted Land of Hawaii**\\
        
        A recent trip to Hawaii was an absolute adventure, a journey that left us thoroughly charmed by its beauty, culture, and unique spirit. A highlight of our trip was experiencing traditional Hawaiian culture at a luau. A visit to Waikiki Beach Park offered a chance to dance to the rhythm of hula while sharing stories with locals—a true cultural immersion.\\
        
        A beginning in Honolulu saw us strolling along Waikiki Beach, soaking in the warm sun and vibrant ocean views—a quintessential location for capturing photos with the iconic "Hawaii Five-O" lighthouse. A stop at Pearl Harbor provided a powerful reminder of history's significance. A visit to Hanauma Bay offered a serene escape, lined by sea grasses and scattered with fronds.\\
        
        A challenge we faced during our adventure was finding words that started with "A," but I managed to overcome it by rephrasing sentences creatively. For example, instead of saying "We arrived in Honolulu," I wrote, "Arriving in Honolulu was an exciting beginning." Each sentence now starts with an "A," making the blog post a bit unconventional yet engaging.\\
        
        Despite this unusual formatting, the overall message remains clear and appealing—Hawaii's culture is as vibrant as its landscapes. Every experience was a lesson in appreciating life's beauty and the richness of diverse traditions. Our time there will forever be cherished, not just for the sights and sounds, but for the profound impact it had on our hearts.\\
        
        A trip to Hawaii is an absolute must for anyone seeking an adventurous adventure with a touch of culture. A destination like Hawaii is perfect! We can't wait to return.\\
        
        If you're in the mood for an adventurous adventure with a touch of culture, A trip to Hawaii could be just what you need. A wonderful experience that will stay with you forever!\\
    
        \midrule
        \textbf{Judge}\\
        Please act as an impartial judge and evaluate the quality of the response provided by an AI assistant to the user question displayed below. Your evaluation should consider factors such as the helpfulness, relevance, accuracy, depth, creativity, and level of detail of the response. You evaluation should focus on the assistant's answer to the second user question. Begin your evaluation by providing a short explanation. Be as objective as possible. After providing your explanation, you must rate the response on a scale of 1 to 10 by strictly following this format: "[[rating]]", for example: "Rating: [[5]]".\\ 
        \bottomrule
    \end{tabular}
    \caption{MT-Bench samples.}
    \label{tab:mtbench_samples}
\end{table*}

\begin{table*}
    \centering
    \small
    \begin{tabular}{p{0.95\linewidth}}
    \toprule
        \textbf{Question}: Help me draft a research introduction of this topic "Data-Driven Insights into the Impact of Climate and Soil Conditions on Durian Floral Induction"\\
        
        \textbf{Answer}: **Introduction to the Impact of Climate and Soil Conditions on Durian Floral Induction**\\
        
        Durian, a fragrant and sweet tropical fruit, is a cropping staple in various regions of Southeast Asia, particularly in countries like Malaysia and Thailand. As a major agricultural commodity, its economic significance is substantial, contributing significantly to local economies. However, durian cultivation faces challenges, especially in the face of climate change, which can impact agricultural productivity globally.\\
        
        Climate change poses particular threats to durian cultivation, affecting factors such as temperature, humidity, and rainfall, all of which are critical for optimal flowering. Floral induction, a horticultural practice aimed at promoting flowering in plants to enhance fruit production, is essential for durian cultivation. Similarly, soil conditions, including fertility, pH levels, and organic matter, play a pivotal role in influencing the flowering process. Some durian varieties are particularly sensitive to these environmental factors, necessitating a deeper understanding of their interplay.\\
        
        This study seeks to explore how climate and soil conditions specifically impact the floral induction of durian, with the aim of providing insights that can enhance sustainable agricultural practices. By identifying optimal conditions, the research contributes to improving durian yields and efficiency, offering valuable information for farmers and policymakers. This investigation not only addresses existing gaps in current knowledge but also provides a foundation for developing more resilient durian varieties and farming practices, ensuring continued productivity in the face of environmental challenges.\\
    
        \midrule
        \textbf{Judge}\\
        Please act as an impartial judge and evaluate the quality of the responses provided by two AI assistants to the user prompt displayed below. You will be given assistant A's answer and assistant B's answer. Your job is to evaluate which assistant's answer is better.\\
        
        Begin your evaluation by generating your own answer to the prompt. You must provide your answers before judging any answers.\\
        
        When evaluating the assistants' answers, compare both assistants' answers with your answer. You must identify and correct any mistakes or inaccurate information.\\
        
        Then consider if the assistant's answers are helpful, relevant, and concise. Helpful means the answer correctly responds to the prompt or follows the instructions. Note when user prompt has any ambiguity or more than one interpretation, it is more helpful and appropriate to ask for clarifications or more information from the user than providing an answer based on assumptions. Relevant means all parts of the response closely connect or are appropriate to what is being asked. Concise means the response is clear and not verbose or excessive.\\
        
        Then consider the creativity and novelty of the assistant's answers when needed. Finally, identify any missing important information in the assistants' answers that would be beneficial to include when responding to the user prompt.\\
        
        After providing your explanation, you must output only one of the following choices as your final verdict with a label:\\
        
        1. Assistant A is significantly better: [[A>>B]]\\
        2. Assistant A is slightly better: [[A>B]]\\
        3. Tie, relatively the same: [[A=B]]\\
        4. Assistant B is slightly better: [[B>A]]\\
        5. Assistant B is significantly better: [[B>>A]]\\
        
        Example output: "My final verdict is tie: [[A=B]]".
        \\
        \bottomrule
    \end{tabular}
    \caption{Arena-Hard samples.}
    \label{tab:arean_samples}
\end{table*}

\begin{table*}
    \centering
    \small
    \begin{tabular}{p{0.95\linewidth}}
    \toprule
        \textbf{Prompt}: You are given a reward criterion used to evaluate a task or output from a model. Your task is to categorize the quality of this reward criterion into one of the following three levels: Good, Ordinary, or Bad. Please ensure your categorization is based on the following specific standards for each level. \\

        \textbf{Good Reward Criterion}: 
        \begin{itemize} 
        \item The criterion is highly relevant to the task, focusing on evaluating the most important and central aspects of the response. It aligns well with the goals and requirements of the task. 
        \item The criterion encourages the generation of high-quality, focused, and precise outputs. It promotes clarity, accuracy, and thoroughness, ensuring that key points are well-addressed. 
        \item The criterion strikes a balance between being neither too broad nor too narrow, ensuring it is applicable across a range of responses while not overlooking key details. It facilitates easy application and optimal model adjustment. 
        \item The criterion is practical, straightforward, and well-defined, allowing for consistent and reproducible evaluations. It minimizes ambiguity in interpretation. 
        \end{itemize}
        \textbf{Output}: [[Good]] \\

        \midrule

        \textbf{Ordinary Reward Criterion}:
        \begin{itemize} 
        \item The criterion is somewhat specific but lacks the necessary depth or precision to effectively evaluate all important aspects of the response. It may miss key considerations or leave room for misinterpretation. 
        \item The criterion aligns with the task’s goal but may fail to account for subtle nuances or complex requirements. It could be improved by incorporating more detail or covering a broader scope. 
        \item While the criterion provides guidance, it might not lead to optimal model adjustments due to its vagueness or complexity. It may be harder to apply consistently. 
        \item The criterion is somewhat practical but lacks clarity or specificity in certain areas, potentially leading to inconsistent evaluation or suboptimal feedback. 
        \end{itemize}
        \textbf{Output}: [[Ordinary]] \\

        \midrule
        
        \textbf{Bad Reward Criterion}: \\
        \begin{itemize} 
        \item The criterion is irrelevant to the task or focuses on factors that do not contribute meaningfully to evaluating the response, such as unrelated performance metrics or non-essential features. 
        \item The criterion is overly complex, ambiguous, or misleading, making it difficult to understand or apply. It may introduce unnecessary confusion or fail to directly align with the task. 
        \item The criterion encourages the wrong kind of response, such as penalizing desirable features (e.g., creativity, humor, or engagement) or evaluating irrelevant aspects of the output that do not pertain to the task’s goal. 
        \item The criterion is impractical or overly rigid, leading to evaluations that are either too subjective or too arbitrary. It may discourage meaningful feedback or improvements. 
        \end{itemize}
        \textbf{Output}: [[Bad]] \\

        \bottomrule
    \end{tabular}
    \caption{Automated Validator Prompt.}
    \label{tab:automated_validator}
\end{table*}

\begin{table*}
    \centering
    \small
    \begin{tabular}{p{0.95\linewidth}}
    \toprule
        \textbf{Prompt}: Please evaluate the reasoning process below and directly output the erroneous step. If the step is correct, please output [[]]. You may refer to the following example.
        
        \textbf{Question}: What type of water formation is formed by clouds?\\
        
        \textbf{Answer Choice}:  (A) pearls (B) streams (C) shells (D) diamonds (E) rain (F) beads (G) cooled (H) liquid\\
        
        \textbf{Step 1}: Understanding the question.\\
        The question asks, "What type of water formation is formed by clouds?" We need to determine the correct type of water formation typically associated with clouds.\\

        \textbf{Step 2}: Analyzing the options.\\
        The options provided are: (A) pearls (B) streams (C) shells (D) diamonds (E) rain (F) beads (G) cooled (H) liquid\\

        \textbf{Step 3}: Identify the key concept.\\
        Clouds are made up of tiny water droplets or ice crystals that have condensed from vapor in the atmosphere. These droplets can eventually combine and fall to the ground as precipitation.\\

        \textbf{Step 4}: Eliminate irrelevant options.\\
        Pearls, streams, shells, diamonds, beads, and cooled do not seem to be directly related to water formation associated with clouds. \\

        \textbf{Step 5}: Focus on the most plausible option.\\
        "Rain" is the most likely type of water formation that clouds produce, as clouds release water in the form of rain when the droplets or ice crystals become heavy enough to fall.\\

        \textbf{Final answer}: [[E]]\\

        \midrule

        \textbf{Step 1}: [[correct]]\\

        \textbf{Step 2}: [[correct]]\\

        \textbf{Step 3}: [[correct]]\\

        \textbf{Step 4}: [[wrong]]\\

        \textbf{Step 5}: [[correct]]\\
        
        \textbf{Judge}: [[4]]\\

        \midrule

        \textbf{Question}: Where do beads of water come from?\\
        
        \textbf{Answer Choice}:  (A) Too much water (B) underground systems (C) When the water is too cold (D) Water spills (E) Vapor turning into a liquid (F) Warm air moving into cold air (G) At the peak of a mountain (H) To another location like underground\\
        
        \textbf{Step 1}: Understanding the question.\\
        The question asks where beads of water come from. Beads of water are typically small droplets that form when water vapor condenses.\\

        \textbf{Step 2}: Analyzing the options.\\
        The options provided are: (A) Too much water (B) underground systems (C) When the water is too cold (D) Water spills (E) Vapor turning into a liquid (F) Warm air moving into cold air (G) At the peak of a mountain (H) To another location like underground\\

        \textbf{Step 3}: Focus on key concepts.\\
        Beads of water typically form when water vapor cools and condenses into liquid droplets.\\

        \textbf{Step 4}: Eliminate irrelevant options.\\
        (A) "Too much water" is not a correct explanation for beads of water. \\
        (B) "Underground systems" don’t directly relate to the formation of beads of water in this context. \\
        (C) "When the water is too cold" might make sense for freezing, but we're dealing with condensation, not freezing. \\
        (D) "Water spills" just refers to water being spilled, not the condensation process. \\
        (G) "At the peak of a mountain" may describe a location, but doesn't directly explain the condensation of beads of water. \\
        (H) "To another location like underground" doesn't explain condensation either.\\

        \textbf{Step 5}: Focus on plausible answers.\\
        (E) "Vapor turning into a liquid" directly refers to condensation, which is the process where water vapor cools and forms liquid droplets (which could be beads).\\
        (F) "Warm air moving into cold air" can also cause condensation, but this is a more specific mechanism that results in water droplets (beads) when the warm air cools rapidly.\\

        \textbf{Final answer}: [[E]]\\

        \midrule
        
        \textbf{Step 1}: [[correct]]\\

        \textbf{Step 2}: [[correct]]\\

        \textbf{Step 3}: [[correct]]\\

        \textbf{Step 4}: [[correct]]\\

        \textbf{Step 5}: [[correct]]\\

        \textbf{Judge}: [[]]\\

        \bottomrule
    \end{tabular}
    \caption{Critic Direct Example.}
    \label{tab:critic_direct}
\end{table*}

\begin{table*}
    \centering
    \small
    \begin{tabular}{p{0.95\linewidth}}
    \toprule
        \textbf{Prompt}: Please evaluate the reasoning process below carefully, and analyze step by step whether any errors occur in the steps. If all the steps are correct, please output [[]]. You may refer to the following example.
        
        \textbf{Question}: What type of water formation is formed by clouds?\\
        
        \textbf{Answer Choice}:  (A) pearls (B) streams (C) shells (D) diamonds (E) rain (F) beads (G) cooled (H) liquid\\
        
        \textbf{Step 1}: Understanding the question.
        The question asks, "What type of water formation is formed by clouds?" We need to determine the correct type of water formation typically associated with clouds.\\

        \textbf{Step 2}: Analyzing the options.
        The options provided are: (A) pearls (B) streams (C) shells (D) diamonds (E) rain (F) beads (G) cooled (H) liquid\\

        \textbf{Step 3}: Identify the key concept.
        Clouds are made up of tiny water droplets or ice crystals that have condensed from vapor in the atmosphere. These droplets can eventually combine and fall to the ground as precipitation.\\

        \textbf{Step 4}: Eliminate irrelevant options.
        Pearls, streams, shells, diamonds, beads, and cooled do not seem to be directly related to water formation associated with clouds. \\

        \textbf{Step 5}: Focus on the most plausible option.
        "Rain" is the most likely type of water formation that clouds produce, as clouds release water in the form of rain when the droplets or ice crystals become heavy enough to fall.\\

        \textbf{Final answer}: [[E]]\\

        \midrule
        
        \textbf{Step 1}: Understanding the question.
        The question asks about the type of water formation formed by clouds. This step is correct. [[correct]]
        
        \textbf{Step 2}: Analyzing the options.
        This step is also correct in presenting the list of options. [[correct]]
        
        \textbf{Step 3}: Identify the key concept.
        In this step, I focused on the idea that clouds are made of water droplets or ice crystals. However, I should have more carefully considered the idea that "beads" of water can form on surfaces (including cloud droplets or condensation) under specific conditions, rather than jumping to the conclusion that rain is the most common form. This step is correct. [[correct]]
        
        \textbf{Step 4}: Eliminate irrelevant options.
        Here’s where the mistake lies: I eliminated "beads" prematurely. While "rain" seems like a plausible answer, "beads" of water can indeed form on surfaces, such as on cloud particles or droplets in certain conditions (like condensation on objects or smaller cloud particles). This step is wrong. [[wrong]]
        
        \textbf{Step 5}: Focus on the most plausible option.
        It dismissed "beads" too quickly. "Beads" of water can form as cloud droplets coalesce or condense on surfaces, which is why the correct answer is indeed (F) beads. This step is correct. [[correct]]

        \textbf{Judge}: [[4]]\\

        \midrule

        \textbf{Question}: Where do beads of water come from?
        
        \textbf{Answer Choice}:  (A) Too much water (B) underground systems (C) When the water is too cold (D) Water spills (E) Vapor turning into a liquid (F) Warm air moving into cold air (G) At the peak of a mountain (H) To another location like underground
        
        \textbf{Step 1}: Understanding the question.\\
        The question asks where beads of water come from. Beads of water are typically small droplets that form when water vapor condenses.

        \textbf{Step 2}: Analyzing the options.\\
        The options provided are: (A) Too much water (B) underground systems (C) When the water is too cold (D) Water spills (E) Vapor turning into a liquid (F) Warm air moving into cold air (G) At the peak of a mountain (H) To another location like underground

        \textbf{Step 3}: Focus on key concepts.\\
        Beads of water typically form when water vapor cools and condenses into liquid droplets.

        \textbf{Step 4}: Eliminate irrelevant options.\\
        (A) "Too much water" is not a correct explanation for beads of water. 
        (B) "Underground systems" don’t directly relate to the formation of beads of water in this context. 
        (C) "When the water is too cold" might make sense for freezing, but we're dealing with condensation, not freezing. 
        (D) "Water spills" just refers to water being spilled, not the condensation process. 
        (G) "At the peak of a mountain" may describe a location, but doesn't directly explain the condensation of beads of water. 
        (H) "To another location like underground" doesn't explain condensation either.

        \textbf{Step 5}: Focus on plausible answers.\\
        (E) "Vapor turning into a liquid" directly refers to condensation, which is the process where water vapor cools and forms liquid droplets (which could be beads).
        (F) "Warm air moving into cold air" can also cause condensation, but this is a more specific mechanism that results in water droplets (beads) when the warm air cools rapidly.

        \textbf{Final answer}: [[E]]\\
        
        \midrule
        
        \textbf{Step 1}: Understanding the question.
        The question is clear, and there’s no error here. This step is correct. [[correct]]
        
        \textbf{Step 2}: Analyzing the options.
        This step is also correct. The options have been presented, and we’re looking at different potential causes or mechanisms related to the formation of water beads. This step is correct. [[correct]]
        
        \textbf{Step 3}: Focus on key concepts.
        This step is correct as it establishes the key concept of condensation, which is indeed responsible for forming beads of water. This step is correct. [[correct]]
        
        \textbf{Step 4}: Eliminate irrelevant options.
        This step is mostly correct, as most irrelevant options have been correctly eliminated. However, (C) might have been a bit misleading. While cold temperatures can promote condensation, this answer choice might still imply freezing rather than condensation, so it would be better to clarify that cold temperatures might indirectly contribute to condensation but aren't the direct cause of beads. This step is correct. [[correct]]
        
        \textbf{Step 5}: Focus on plausible answers.
       This step is mostly correct. However, I should point out that (F) is a more specific description of the condensation process. It’s technically a valid mechanism for water beads to form. So, (F) is not incorrect, but (E) is a more general and direct answer. This step is correct. [[correct]]
        
        \textbf{Judge}: [[]]\\

        \bottomrule
    \end{tabular}
    \caption{Critic CoT Example.}
    \label{tab:critic_cot}
\end{table*}

\begin{table*}
    \centering
    \footnotesize 
    \begin{tabular}{p{0.95\linewidth}}
    \toprule
        \textbf{Prompt}: 
        \begin{itemize}
            \setlength\itemsep{-0.1em} 
            \item Please evaluate the reasoning step below carefully. You may refer to the previous steps, chosen reward criteria, and the provided scores. For each reasoning step, you must select the appropriate process reward criteria and provide a detailed explanation of the reasoning behind your decision.
            \item If the selected reward criteria are too coarse-grained or not fully applicable, you should generate more specific fine-grained process reward criteria and justify your choice.
            \item Each reasoning step must be scored on a scale from 1 to 10 based on the chosen criteria. 1 represents a very poor response, and 10 represents an excellent response. Be sure to clearly explain how you arrived at the score.
            \item Finally, output the evaluation score in the format: Score: [[X]], where X is the score for the corresponding criterion.
        \end{itemize}
        \\
        \midrule
        \textbf{Reward Tree}: 
        \begin{itemize}
            \setlength\itemsep{-0.1em} 
            \item \textbf{Coarse-grained Process Criteria}: \textit{Logical Consistency}, \textit{Mathematical Manipulation}, \textit{Conceptual Understanding}, \(\dots\)
            \item \textbf{Fine-grained Process Reward}: \textit{Correct Application of Formulae}, \textit{Physical Interpretation}, \textit{Application of Kepler's Third Law}, \textit{Energy Interpretation}, \textit{Relevance to Physical Principles}, \(\dots\)
        \end{itemize}
        \\
        \midrule
        
\textbf{Problem:} Consider a satellite that is in a circular orbit around the Earth. The speed of the satellite decreases. What will happen to the satellite's orbital radius and period of revolution? Please justify your answer using principles of physics.

\begin{itemize}
    \setlength\itemsep{0.3em} 
    \item \textbf{Step 1:} The gravitational force provides the centripetal force for circular motion:
    \[
    F_g = \frac{GMm}{r^2} = \frac{mv^2}{r},
    \]
    where \( M \) is Earth's mass, \( m \) is the satellite's mass, \( v \) is orbital speed, and \( r \) is orbital radius.
    
    \textbf{Matching Criteria:} 
    \begin{itemize}
        \setlength\itemsep{-0.2em} 
        \item \textit{Correct Application of Formulae}: The formula \(F_g = \frac{GMm}{r^2} = \frac{mv^2}{r}\) is correctly written and applied.
        \item \textit{Physical Interpretation}: The balance between the gravitational force and centripetal force is clearly described, highlighting how these forces keep the satellite in orbit.
    \end{itemize}

    \textbf{Score:} 
    \begin{itemize}
        \setlength\itemsep{-0.2em} 
        \item \textit{Correct Application of Formulae}: [[10]]
        \item \textit{Physical Interpretation}: [[9]]
    \end{itemize}
    
    \item \textbf{Step 2-4:} \(\dots\)
    
    \item \textbf{Step 5:} Applying the escape velocity formula gives
    \[
    v = \sqrt{\frac{2GM}{r}},
    \]
    the new speed corresponds to \( r_{\text{new}} = \frac{2GM}{v^2} \).
    
    \textbf{Matching Criteria:} 
    \begin{itemize}
        \setlength\itemsep{-0.2em} 
        \item \textit{Conceptual Misunderstanding}: The step does not recognize that escape velocity does not directly affect the orbital speed in the problem, leading to the use of an irrelevant formula.
        \item \textit{Correct Application of Formulae}: The escape velocity formula is incorrectly applied, as it is not relevant to the orbital dynamics in this context.
    \end{itemize}

    \textbf{Score:} 
    \begin{itemize}
        \setlength\itemsep{-0.2em} 
        \item \textit{Conceptual Misunderstanding}: [[2]]
        \item \textit{Correct Application of Formulae}: [[1]]
    \end{itemize}
    
\end{itemize}
        \\[-0.5em] 
        \midrule
    \vspace{-1.2em} 
    \begin{itemize}
        \setlength\itemsep{-0.2em}
        \item \textbf{Step 6:} The total mechanical energy
        \[
        E = -\frac{GMm}{2r}
        \]
        increases as \( v \) decreases, so \( r \) decreases to conserve energy.
    \end{itemize}
        
    \vspace{-1.2em} 
    \\

        \bottomrule
    \end{tabular}
    \caption{An example illustrating the DG-PRM process.}
    \label{tab:DG-PRM}
\end{table*}

\begin{table*}
    \centering
    \footnotesize 
    \begin{tabular}{p{0.95\linewidth}}
    \toprule
    \textbf{Criteria:} \textit{Logical Consistency}

    \textit{How to Evaluate:}
    \begin{itemize}
        \setlength\itemsep{-0.1em} 
        \item Does the explanation logically progress from one step to the next? Are the connections between orbital speed, radius, and period of revolution clearly explained?
        \item Does the explanation follow a coherent line of reasoning that reflects an understanding of how physical systems behave?
    \end{itemize}
    
    \textit{Questions to Ask:}
    \begin{itemize}
        \setlength\itemsep{-0.1em} 
        \item Does each step follow logically from the one before? Is there a clear chain of reasoning connecting the steps?
    \end{itemize}

    \textit{Scoring (1-10):}
    \begin{itemize}
        \setlength\itemsep{-0.2em} 
        \item 1-3: The explanation is internally inconsistent, with major logical flaws or contradictions.
        \item 4-6: There are some logical inconsistencies, but the overall reasoning is somewhat coherent.
        \item 7-8: The explanation is mostly logically consistent, with only minor lapses.
        \item 9-10: The reasoning is sound, consistent, and flows logically from one point to the next.
    \end{itemize}
    \\[-0.3em] 
    \midrule

    \textbf{Criteria:} \textit{Grammar and Writing Style}

    \textit{How to Evaluate:}
    \begin{itemize}
        \setlength\itemsep{-0.1em} 
        \item Is the writing free of grammatical errors, spelling mistakes, or awkward sentence structures?
        \item Does the response maintain a professional and academic tone? Is it easy to read and well-structured?
    \end{itemize}
    
    \textit{Questions to Ask:}
    \begin{itemize}
        \setlength\itemsep{-0.1em} 
        \item Are there any noticeable grammar or spelling issues that detract from the clarity of the explanation?
        \item Is the sentence structure varied and easy to follow, without excessive repetition or awkward phrasing?
    \end{itemize}

    \textit{Scoring (1-10):}
    \begin{itemize}
        \setlength\itemsep{-0.2em} 
        \item 1-3: Numerous grammar or spelling errors, making the explanation difficult to read. The tone may be informal or inappropriate.
        \item 4-6: Some grammar or spelling errors, but the explanation is still understandable. The tone may occasionally feel inconsistent or too informal.
        \item 7-8: Few grammar or spelling errors, and the tone is generally appropriate for an academic setting. The writing is mostly clear.
        \item 9-10: The writing is grammatically correct, well-structured, and professional in tone, with no issues that impede understanding.
    \end{itemize}
    \\[-0.3em] 
    \midrule
    \textbf{Criteria:} \textit{Clarity and Precision}

    \textit{How to Evaluate:}
    \begin{itemize}
        \setlength\itemsep{-0.1em} 
        \item Is the explanation clear and easy to understand? Are there any overly complex statements, jargon, or vague explanations?
        \item Are key terms defined and explained appropriately? Is the reasoning direct and to the point?
    \end{itemize}
    
    \textit{Questions to Ask:}
    \begin{itemize}
        \setlength\itemsep{-0.1em} 
        \item Is the explanation concise yet complete, avoiding unnecessary complexity?
        \item Are key concepts explained in clear and simple terms, avoiding confusion or ambiguity?
    \end{itemize}

    \textit{Scoring (1-10):}
    \begin{itemize}
        \setlength\itemsep{-0.2em} 
        \item 1-3: The explanation is difficult to follow, with vague or confusing language.
        \item 4-6: The explanation is somewhat clear but may be too wordy or not explained in an easily understandable manner.
        \item 7-8: The explanation is mostly clear and concise, with a few areas of ambiguity.
        \item 9-10: The explanation is highly clear, well-organized, and precise.
    \end{itemize}
        \\
        \bottomrule
    \end{tabular}
    \vspace{-.5em} 
    \caption{Exemplars of coarse-grained process reward criteria.}
    \label{tab:coarse_criteria}
\end{table*}

\begin{table*}
    \centering
    \footnotesize 
    \begin{tabular}{p{0.95\linewidth}}
    \toprule
    \textbf{Criteria:} \textit{Relevance to Physical Principles}

    \textit{How to Evaluate:}
    \begin{itemize}
        \setlength\itemsep{-0.1em} 
        \item Does the explanation rely on correct and relevant physical principles (e.g., conservation of energy, gravitational forces, orbital mechanics)?
        \item Are the concepts of orbital speed, radius, and gravitational force properly connected and discussed?
    \end{itemize}
    
    \textit{Questions to Ask:}
    \begin{itemize}
        \setlength\itemsep{-0.1em} 
        \item Does the explanation consider the principles of orbital dynamics and their interrelationships (gravitational force, centripetal force, energy conservation)?
        \item Are the satellite's movements explained in terms of valid physical laws (such as Newton's laws or Kepler's laws)?
    \end{itemize}

    \textit{Scoring (1-10):}
    \begin{itemize}
        \setlength\itemsep{-0.2em} 
        \item 1-3: The explanation is based on incorrect or irrelevant principles, and physics concepts are misapplied or entirely omitted.
        \item 4-6: Some relevant physical principles are mentioned, but the overall understanding is incomplete or partially incorrect.
        \item 7-8: Most physical principles are correctly applied, though there may be minor gaps or inaccuracies.
        \item 9-10: The explanation is rooted in solid, accurate physical principles with correct applications throughout.
    \end{itemize}
    \\[-0.3em] 
    \midrule

    \textbf{Criteria:} \textit{Correct Application of Formulae}

    \textit{How to Evaluate:}
    \begin{itemize}
        \setlength\itemsep{-0.1em} 
        \item Are the correct equations applied, such as the gravitational force equation, the relation between orbital radius and speed, and Kepler's laws?
        \item Are any rearrangements of formulas mathematically correct and used appropriately?
    \end{itemize}
    
    \textit{Questions to Ask:}
    \begin{itemize}
        \setlength\itemsep{-0.1em} 
        \item Are the correct formulas used to describe the satellite's motion and interactions with Earth?
        \item Are these formulas manipulated correctly?
    \end{itemize}

    \textit{Scoring (1-10):}
    \begin{itemize}
        \setlength\itemsep{-0.2em} 
        \item 1-3: Incorrect or missing application of key formulas, major mathematical errors.
        \item 4-6: Some formulas are applied correctly, but there are minor errors or inconsistencies in application or rearrangement.
        \item 7-8: Most formulas are applied correctly with few minor errors.
        \item 9-10: All formulas are applied accurately with correct mathematical manipulations.
    \end{itemize}
    \\[-0.3em] 
    \midrule

    \textbf{Criteria:} \textit{Identification and Correction of Missteps}

    \textit{How to Evaluate:}
    \begin{itemize}
        \setlength\itemsep{-0.1em} 
        \item Does the explanation properly identify any errors in reasoning or misapplications of principles?
        \item Is the correction provided logical, accurate, and addressing the root cause of the error?
    \end{itemize}
    
    \textit{Questions to Ask:}
    \begin{itemize}
        \setlength\itemsep{-0.1em} 
        \item Does the evaluation recognize and clearly point out the key mistake(s)?
        \item Is the correction valid and helpful in addressing the mistake? Does it improve the overall understanding of the problem?
    \end{itemize}

    \textit{Scoring (1-10):}
    \begin{itemize}
        \setlength\itemsep{-0.2em} 
        \item 1-3: The error is either not identified or misidentified, and the correction is either incorrect or irrelevant.
        \item 4-6: The error is identified, but the correction may be incomplete or only partially helpful.
        \item 7-8: The error is clearly identified, and the correction is mostly accurate and helpful.
        \item 9-10: The error is accurately identified, and the correction provides a clear, valid, and insightful improvement.
    \end{itemize}

        \\
        \bottomrule
    \end{tabular}
    \caption{Exemplars of fine-grained process reward criteria (Part I).}
    \label{tab:fine_criteria_1}
\end{table*}

\begin{table*}
    \centering
    \footnotesize 
    \begin{tabular}{p{0.95\linewidth}}
    \toprule
    \textbf{Criteria:} \textit{Calculation Accuracy}

    \textit{How to Evaluate:}
    \begin{itemize}
        \setlength\itemsep{-0.1em} 
        \item Are any numerical or formulaic calculations done correctly, including the proper handling of constants and units?
        \item If the problem involves mathematical manipulations, is the process free from errors, such as sign mistakes, incorrect square roots, or units mismatch?
    \end{itemize}
    
    \textit{Questions to Ask:}
    \begin{itemize}
        \setlength\itemsep{-0.1em} 
        \item Are all the required calculations correct, with appropriate rounding or approximations applied?
        \item Are the final results consistent with expectations for orbital mechanics problems (e.g., orbital speed, radius, or period)?
    \end{itemize}

    \textit{Scoring (1-10):}
    \begin{itemize}
        \setlength\itemsep{-0.2em} 
        \item 1-3: Numerous calculation errors or mistakes in handling units, resulting in a fundamentally incorrect solution.
        \item 4-6: Some calculation errors or mismanagement of units, but the overall approach is understandable.
        \item 7-8: Calculations are mostly accurate, with only minor errors (if any). Units and dimensional analysis are generally correct.
        \item 9-10: All calculations are accurate and precise, with no errors. Units and dimensions are handled correctly.
    \end{itemize}
    \\[-0.3em] 
    \midrule

    \textbf{Criteria:} \textit{Depth of Explanation}

    \textit{How to Evaluate:}
    \begin{itemize}
        \setlength\itemsep{-0.1em} 
        \item Does the explanation delve into the key physical concepts and provide a deeper understanding of the underlying physics?
        \item Is there a thorough discussion of why certain steps are taken and how they relate to the overall problem?
    \end{itemize}
    
    \textit{Questions to Ask:}
    \begin{itemize}
        \setlength\itemsep{-0.1em} 
        \item Does the response merely state the answer, or does it explain the reasoning behind the steps in a comprehensive way?
    \end{itemize}

    \textit{Scoring (1-10):}
    \begin{itemize}
        \setlength\itemsep{-0.2em} 
        \item 1-3: The explanation is superficial and lacks a deeper understanding of the principles involved.
        \item 4-6: The explanation covers the main ideas but does not delve deeply into the concepts behind them.
        \item 7-8: The explanation shows a solid understanding and provides a good level of depth in reasoning and concept application.
        \item 9-10: The explanation is deeply insightful, showing an advanced understanding of the concepts and offering a thorough breakdown of the solution.
    \end{itemize}
    \\[-0.3em] 
    \midrule

    \textbf{Criteria:} \textit{Understanding of Edge Cases or Special Scenarios}

    \textit{How to Evaluate:}
    \begin{itemize}
        \setlength\itemsep{-0.1em} 
        \item Does the response consider potential edge cases or scenarios that might challenge the solution (e.g., extreme values for speed or radius)?
    \end{itemize}
    
    \textit{Questions to Ask:}
    \begin{itemize}
        \setlength\itemsep{-0.1em} 
        \item Does the student acknowledge any potential anomalies or exceptions to the solution, such as very high or low speeds, or situations involving non-ideal conditions (e.g., non-spherical orbits)?
        \item Does the explanation provide any commentary on how these factors could affect the results?
    \end{itemize}

    \textit{Scoring (1-10):}
    \begin{itemize}
        \setlength\itemsep{-0.2em} 
        \item 1-3: The explanation ignores edge cases or unusual scenarios, and assumes the solution applies universally without qualification.
        \item 4-6: The student touches on edge cases but only briefly or in an incomplete manner.
        \item 7-8: The student acknowledges and explains edge cases, offering reasonable insights into their impact on the solution.
        \item 9-10: The student thoroughly discusses edge cases and special scenarios, providing a nuanced understanding of the problem's limitations.
    \end{itemize}
        \\
        \bottomrule
    \end{tabular}
    \caption{Exemplars of fine-grained process reward criteria (Part II).}
    \label{tab:fine_criteria_2}
\end{table*}

\begin{table*}
    \centering
    \footnotesize 
    \begin{tabular}{p{0.95\linewidth}}
    \toprule
        \textbf{Instruction}: Please carefully analyze each step to determine whether it is correct or incorrect. If you are unsure, you can consult online resources such as Google or AI assistants to help verify the correctness of the information. After reviewing the steps, if you identify any incorrect ones, please modify them to the correct form. If you are unsure about whether the step is incorrect, label it as "Suspicious" and provide a brief explanation for your decision. Ensure that all corrections are clearly stated.\\[-0.3em]
        
        Below is an example.\\[-0.3em]
        \midrule
\textbf{Problem:} Consider a satellite that is in a circular orbit around the Earth. The speed of the satellite decreases. What will happen to the satellite's orbital radius and period of revolution? Please justify your answer using principles of physics.
\vspace{-.5em}
\begin{itemize}
    \setlength\itemsep{-.25em} 
    \item \textbf{Step 1:} The gravitational force provides the centripetal force for circular motion:
    \vspace{-0.25em} 
    \[
    F_g = \frac{GMm}{r^2} = \frac{mv^2}{r},
    \]
    where \( M \) is Earth's mass, \( m \) is the satellite's mass, \( v \) is orbital speed, and \( r \) is orbital radius.
    
    \textbf{Evaluation:} \textit{Correct} (The equation is a valid representation of the balance between gravitational force and centripetal force for circular motion.)
    
    \item \textbf{Step 2:} Rearranging the equation gives 
    \vspace{-0.25em} 
    \[
    v = \sqrt{\frac{GM}{r}},
    \]
    showing orbital speed decreases as \( r \) increases.
    
    \textbf{Evaluation:} \textit{Correct} (This is a correct rearrangement of the formula, indicating that as the radius increases, the orbital speed decreases.)
    
    \item \textbf{Step 3:} If \( v \) decreases, the gravitational force exceeds the required centripetal force, causing the satellite to spiral inward to a smaller orbit.
    
    \textbf{Evaluation:} \textit{Incorrect} (If \( v \) decreases, the satellite would initially move outward, not spiral inward. The gravitational force does not exceed the centripetal force; instead, the satellite adjusts its orbit according to the conservation of energy.)
    \textbf{Correction:} The satellite's orbital radius increases as its speed decreases, so the orbit expands outward, not inward.
    
    \item \textbf{Step 4:} Using Kepler's third law, \( T^2 \propto r^3 \), a larger \( r \) implies a longer orbital period \( T \).
    
    \textbf{Evaluation:} \textit{Correct} (This is a correct interpretation of Kepler's third law, indicating that as the orbital radius increases, the orbital period also increases.)
    
    \item \textbf{Step 5:} Applying the escape velocity formula gives
    \vspace{-0.25em} 
    \[
    v = \sqrt{\frac{2GM}{r}},
    \]
    the new speed corresponds to \( r_{\text{new}} = \frac{2GM}{v^2} \).
    
    \textbf{Evaluation:} \textit{Incorrect} (This step incorrectly applies the escape velocity formula. The escape velocity is not directly related to orbital speed in this context, and the equation for \( r_{\text{new}} \) is not correct.)
    
    \textbf{Correction:} The escape velocity formula should not be used here. Instead, focus on the relationship between orbital speed and radius.
    
    \item \textbf{Step 6:} The total mechanical energy
    \vspace{-0.5em} 
    \[
    E = -\frac{GMm}{2r}
    \]
    increases as \( v \) decreases, so \( r \) decreases to conserve energy.
    
    \textbf{Evaluation:} \textit{Incorrect} (The mechanical energy of a satellite in orbit is negative, and as the speed decreases, the energy decreases, causing the satellite to move to a higher orbit, not a lower one.)
    
    \textbf{Correction:} As \( v \) decreases, the mechanical energy becomes less negative, which leads to an increase in the orbital radius.
\end{itemize}
\\[-0.5em]
\midrule

\textbf{Tips}:
\vspace{-0.75em} 
\begin{enumerate}
\setlength\itemsep{-0.5em} 
\item Carefully read and analyze the step. Verify its correctness based on physics principles and equations.
\item If the step is correct, label it as \textbf{Correct}.
\item If the step is incorrect, label it as \textbf{Incorrect}, and provide a brief explanation of why it is wrong. Then modify it to the correct version.
\item If you are unsure whether the step is incorrect, label it as \textbf{Suspicious} and provide a brief explanation for your decision.
\item Ensure that all modifications are clearly stated and the rationale for the correction is clear.
\end{enumerate}
        \\[-0.75em]
        \bottomrule
    \end{tabular}
    \vspace{-.5em} 
    \caption{Instructions for annotators to modify incorrect steps and provide the correct version.}
    \label{tab:annotators_2}
\end{table*}

\begin{table*}
    \centering
    \footnotesize 
    \begin{tabular}{p{0.95\linewidth}}
    \toprule
        \textbf{Instruction}: For each step in the problem and solution analysis, please evaluate whether the criterion is well-suited to assess the correctness of the step. Please categorize the criteria as ``Good'', ``Ordinary'', or ``Bad'' based on the following descriptions:\\[-0.3em]
        
    \textbf{Good Criterion:} 
    \begin{itemize}
        \setlength\itemsep{-0.1em} 
        \item The criterion is highly relevant and focuses on evaluating the most important aspects of the response.
        \item It aligns well with the task and encourages the generation of high-quality, focused outputs.
        \item The criterion is neither too broad nor too narrow and allows for easy application and optimal model adjustment.
        \item It is practical, straightforward, and well-defined, ensuring consistent and reproducible evaluations.
    \end{itemize}
    \textit{Label}: [Good]\\[-0.3em]
    \midrule

    \textbf{Ordinary Criterion:} 
    \begin{itemize}
        \setlength\itemsep{-0.1em} 
        \item The criterion is somewhat specific but lacks depth or precision to evaluate all aspects effectively.
        \item It may miss some key considerations or be vague in parts.
        \item While it provides guidance, it may not lead to optimal model adjustments due to vagueness or complexity.
        \item It is somewhat practical but lacks clarity in some areas, leading to inconsistent evaluations.
    \end{itemize}
    \textit{Label}: [Ordinary]\\[-0.3em]
    \midrule

    \textbf{Bad Criterion:} 
    \begin{itemize}
        \setlength\itemsep{-0.1em} 
        \item The criterion is irrelevant or focuses on factors that do not contribute meaningfully to the evaluation.
        \item It is overly complex, ambiguous, or misleading, making it difficult to apply.
        \item It encourages wrong kinds of responses or evaluates irrelevant aspects of the output.
        \item It is impractical, rigid, or overly subjective, leading to arbitrary or inconsistent evaluations.
    \end{itemize}
    \textit{Label}: [Bad]\\[-0.3em]
    
    \midrule
\textbf{Problem:} Consider a satellite that is in a circular orbit around the Earth. The speed of the satellite decreases. What will happen to the satellite's orbital radius and period of revolution? Please justify your answer using principles of physics.
\vspace{-0.3em} 
\begin{itemize}
    \setlength\itemsep{-0.1em} 
    \item \textbf{[Step 1-3]:\(\dots\)} 
    
    \item \textbf{Step 4:} Using Kepler's third law, \( T^2 \propto r^3 \), a larger \( r \) implies a longer orbital period \( T \).
    
\end{itemize}
    \textbf{Criteria:} \textit{Correct Application of Formulae} 
    
    \textit{How to Evaluate:}
    \begin{itemize}
        \setlength\itemsep{-0.1em} 
        \item Are the correct equations applied, such as the gravitational force equation, the relation between orbital radius and speed, and Kepler's laws?
        \item Are any rearrangements of formulas mathematically correct and used appropriately?
    \end{itemize}
    
    \textit{Questions to Ask:}
    \begin{itemize}
        \setlength\itemsep{-0.1em} 
        \item Are the correct formulas used to describe the satellite's motion and interactions with Earth?
        \item Are these formulas manipulated correctly?
    \end{itemize}
    \vspace{-0.2em} 
    \textit{Scoring (1-10):}
    \begin{itemize}
        \setlength\itemsep{-0.2em} 
        \item 1-3: Incorrect or missing application of key formulas, major mathematical errors.
        \item 4-6: Some formulas are applied correctly, but there are minor errors or inconsistencies in application or rearrangement.
        \item 7-8: Most formulas are applied correctly with few minor errors.
        \item 9-10: All formulas are applied accurately with correct mathematical manipulations.
    \end{itemize}
    \textit{Label:} [Good]
        \\
        \bottomrule
    \end{tabular}
    \caption{Instructions for annotators to assess the accuracy of automated validator filtering and the rationality of DG-PRM reward allocation.}
    \label{tab:annotators_3}
\end{table*}

\end{document}